\documentclass[runningheads]{llncs}
\usepackage[T1]{fontenc}
\usepackage{graphicx}
\usepackage{subcaption}
\usepackage{colortbl}
\usepackage{booktabs}
\usepackage{xspace}
\usepackage[misc]{ifsym}
\usepackage[numbers]{natbib}

\usepackage{amsmath,amsfonts,bm}

\def\eqref#1{equation~\ref{#1}}

\def\1{\bm{1}}

\DeclareMathAlphabet{\mathsfit}{\encodingdefault}{\sfdefault}{m}{sl}
\SetMathAlphabet{\mathsfit}{bold}{\encodingdefault}{\sfdefault}{bx}{n}

\DeclareMathOperator*{\argmax}{arg\,max}

\usepackage{floatrow}
\usepackage{makecell, cellspace, caption}
\usepackage{algorithm,algpseudocode}
\usepackage{amsmath}
\usepackage{booktabs}
\usepackage{multirow}
\usepackage{adjustbox}
\usepackage{rotating}
\usepackage{comment}
\usepackage{xcolor}         
\usepackage{bm}
\usepackage{graphicx}
\definecolor{cvprblue}{rgb}{0.21,0.49,0.74}
\usepackage[pagebackref,breaklinks,colorlinks,allcolors=cvprblue]{hyperref}

\def\etal{{et al.\xspace}}
\newcommand{\norm}[1]{\left\lVert#1\right\rVert}

\newcommand{\tsub}[1]{\textsubscript{#1}}

\newcommand{\myfirstpara}[1]{\par \noindent \textbf{{#1}.}}
\newcommand{\mypara}[1]{\vspace{0.2em} \myfirstpara{#1}}
\newcommand{\vlms}{\texttt{VLMs}\xspace}
\newcommand{\llm}{\texttt{LLM}\xspace}
\newcommand{\myclip}{\texttt{CLIP}\xspace}
\newcommand{\tpt}{\texttt{TPT}\xspace}
\newcommand{\ctpt}{\texttt{C-TPT}\xspace}
\newcommand{\vtpt}{\texttt{VTPT}\xspace}
\newcommand{\ece}{\texttt{ECE}\xspace}
\newcommand{\acc}{\texttt{Acc}\xspace}

\newcommand{\mtas}{\texttt{MTAS}\xspace}
\newcommand{\atfd}{\texttt{ATFD}\xspace}

\newcommand{\tca}{\texttt{TCA}\xspace}
\newcommand{\ood}{\texttt{OOD}\xspace}
\newcommand{\llmtt}{\texttt{LLM}\xspace}
\newcommand{\llms}{\texttt{LLMs}\xspace}
\newcommand{\gpt}{\texttt{GPT4}\xspace}

\newcommand{\bE}{\mathbb{E}}

\newcommand{\cL}{\mathcal{L}}

\newcommand{\cX}{\mathcal{X}}
\newcommand{\cY}{\mathcal{Y}}
\newcommand{\cD}{\mathcal{D}}
\newcommand{\yh}{\widehat{y}}
\newcommand{\ph}{\widehat{p}}
\newcommand{\pb}{\textbf{p}\xspace}
\newcommand{\pbar}{\bar{p}}
\newcommand{\tbar}{\bar{t}}
\newcommand{\tdbar}{\bar{\bar{t}}}

\newcommand{\eg}{e.g.\xspace}
\usepackage{mwe}
\usepackage{enumerate}
\usepackage[inline]{enumitem}
\usepackage[capitalize]{cleveref} 
\usepackage{wrapfig}

\begin{document}
\title{Prompting without Panic: Attribute-aware, Zero-shot, Test-Time Calibration}

\titlerunning{Prompting without Panic: Attribute-aware, Zero-shot, Test-Time Calibration}

\author{Ramya Hebbalaguppe$^{1,2}$\textsuperscript{\textsection}  \quad
	Tamoghno Kandar$^{2}$\textsuperscript{\textsection}  \quad
	\quad Abhinav Nagpal$^1$ \quad
    \quad Chetan Arora$^1$ \\
	$^1$IIT Delhi  \quad$^2$ TCS Research Labs \\ \textbf{Project Webpage:} \url{https://promptwithoutpanic.github.io/}
}
\institute{}
\authorrunning{Hebbalaguppe et al.}
\begingroup\renewcommand\thefootnote{\textsection}
\footnotetext{Equal contribution}
\endgroup
\maketitle              
\begin{abstract}
Vision language models (\vlms) have become effective tools for image recognition, primarily due to their self-supervised training on large datasets. Their performance can be enhanced further through test-time prompt tuning (\tpt). However, \tpt's singular focus on accuracy improvement often leads to a decline in confidence calibration, restricting its use in safety-critical applications. In this work, we make two contributions: 
\begin{enumerate*}[label=\textbf{(\arabic*)}]
\item We posit that random or naive initialization of prompts leads to overfitting on a particular test sample, and is one of the reasons for miscalibration of \vlms after \tpt. To mitigate the problem, we propose careful initialization of test time prompt using prior knowledge about the target label attributes from a large language model (\llm). 
\item We propose a novel regularization technique to preserve prompt calibration during test-time prompt tuning (\tpt). This method simultaneously minimizes intraclass distances while maximizing interclass distances between learned prompts.
\end{enumerate*}
Our approach achieves significant calibration improvements across multiple CLIP architectures and 15 diverse datasets, demonstrating its effectiveness for \tpt. We report an average expected calibration error (\ece) of 4.11 with our method, TCA, compared to 11.7 for vanilla \tpt\cite{tpt}, 6.12 for C-TPT\cite{yoon2024ctpt} (ICLR'24), 6.78 for DiffTPT\cite{feng2023diverse} (CVPR'23), and 8.43 for PromptAlign\cite{abdul2024align} (NeurIPS'23). The code is publicly accessible at \url{https://github.com/rhebbalaguppe/TCA_PromptWithoutPanic}.
\end{abstract}

\section{Introduction}
\label{sec:intro}

\begin{figure}[t]
    \includegraphics[width=\linewidth,height=0.35\linewidth]{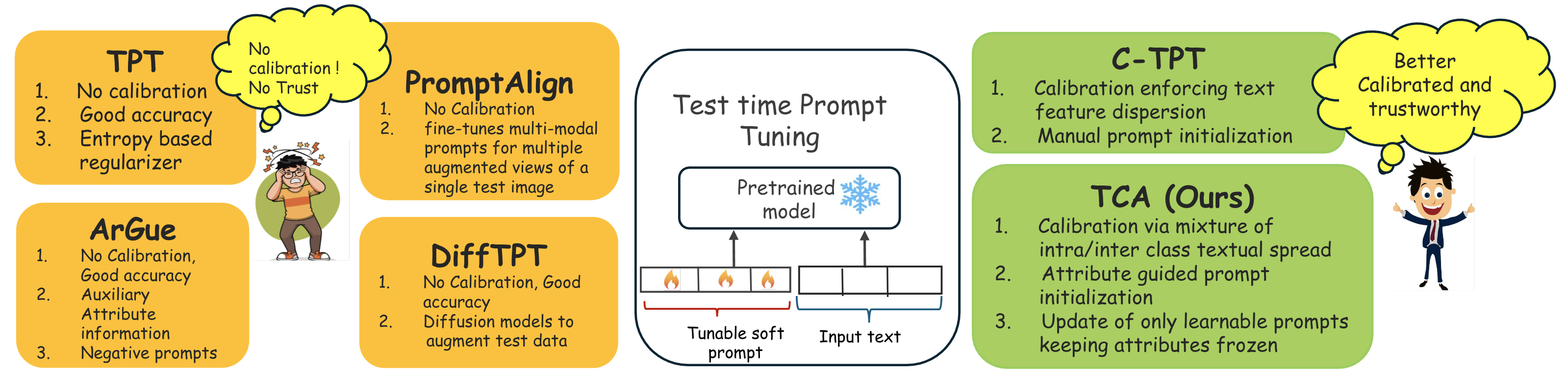}
    \caption{\textbf{Conceptual comparison between our proposed \tca vs. the contemporaries.} Test-time prompt tuning methods, such as \tpt \cite{tpt}, learn test-time prompts through parameter optimization. However, these methods often face performance disadvantages in calibration, as they struggle to dynamically adapt to varying textual feature distributions, limiting effective prompt calibration. Methods, \texttt{ArgGue}\cite{tian2024argue}, \texttt{DiffTPT}\cite{feng2023diverse}, and \texttt{PromptAlign}\cite{abdul2024align} do not explicitly optimize for calibration. Although \ctpt \cite{yoon2024ctpt} introduces enhancements in calibration, it still falls short in capturing \emph{nuanced visual attributes} that contribute to precise prompt conditioning leading to suboptimal prompt specificity. Our method termed \textbf{T}est-time \textbf{C}alibration via \textbf{A}ttribute Alignment (\texttt{TCA}) infuses relevant attribute information providing context via \llms and captures intra/inter-class textual attribute spread improving prompt calibration. \textbf{Note:} \tca works in zero-shot and test-time settings without any labeled data, making it very practical for real-world deployment where data annotation is infeasible. No model finetuning required: Only prompts are updated at test time; base vision and text encoders are kept frozen.}
    \label{fig:conceptualDiff}
\end{figure}
\myfirstpara{\vlms and Confidence calibration}
Vision-Language Models (\vlms) have unlocked transformative applications across a wide range of fields, from healthcare diagnostics \cite{vlmhealthcare} to assistive solutions for visually impaired \cite{zhao2024vialm}. However, recent findings 
\cite{tu2024empiricalstudymatterscalibrating} reveal that \vlms suffer from miscalibration, which can hinder model trustworthiness in critical applications. Traditional calibration methods rely on large labeled datasets, posing significant limitations for settings like test-time adaptation, where the labeled data is unavailable or infeasible to obtain. Inspired by the success of \vlms in generalizing to unseen data in a zero-shot setting \cite{yoon2024ctpt}, in this paper we focus on zero-shot setting, and adapt these models using prompt tuning.

\myfirstpara{Prompt Tuning}
Test-time prompt tuning (\tpt) has emerged as a promising approach to improve generalization of \vlms, offering a way to adapt prompts to specific contexts without requiring any labeled data from the target domain. Hard prompts \cite{tta1}, often composed of fixed vocabulary tokens from standard templates like \texttt{``A photo of a \{class name\}"} can simplify prompt creation. However, \cite{yoon2024ctpt} indicate that more flexible prompt designs, such as soft prompts or learned embeddings, can significantly enhance a model's adaptability and effectiveness. On the other hand, domain-specific prompt creation for image-text models requires substantial expertise and time, with no guarantee of optimal results despite extensive engineering efforts\cite{ghosal2024intcoopinterpretabilityawarevisionlanguageprompt}. Shu \etal \cite{tpt} suggested a \tpt technique (hereinafter referred to as Vanilla TPT (\vtpt)) which aims to enhance the accuracy of \myclip based models by minimizing the entropy in the prediction distribution as a self-supervision signal during test time. However, a reduction in entropy leads the model to generate overconfident predictions, a characteristic often observed in models trained with cross-entropy loss \cite{calibrationmodern,yoon2024ctpt}. Fig. \ref{fig:conceptualDiff} illustrates the conceptual distinction between existing prompt tuning approaches and the method proposed in this work.


\mypara{Contributions}
This work focuses on \tpt strategy to improve model's calibration. At first, this may seem infeasible since various calibration techniques employed in standard supervised training of neural networks require substantial amounts of labeled training data, which restricts their applicability in test-time prompt tuning scenarios for \myclip based models. Here, we come up with a clever workaround, by extracting label attributes using a \llm, and leveraging them in \tpt instead of label supervision.

\begin{enumerate}[leftmargin=*]
\item \textbf{Attribute-Aware Prompting for Improved Calibration:} 
Unlike the contemporary methods that directly attach soft prompts before class names, we append the model with precise visual attributes produced by an \llm that provide rich context. The visual attributes are sorted by their \textbf{relevance}. It may be noted that a particular attribute may be relevant for more than one labels. Hence, by aligning the visual embeddings with the chosen attributes allows a model to not only demonstrate that it recognizes features that are crucial for distinguishing the correct class from others, but also allows the model to express its prediction uncertainty in terms of the ambiguous attributes. Multiple relevant attributes also enhance the compositional nature of visual data as they serve as semantic anchors. Their incorporation in soft prompt design improves image-text alignment scores as they establish interpretable correspondences between visual and linguistic embeddings.
\item \textbf{Regularization Loss:} Proposed visual attributes-based prompt initialization allows the model a much better starting point compared to random initialization and prevents overfitting in the presence of limited variations in the single sample (and its augmentation) based training. However, the gradient-based update of the prompts may still overfit the prompts to the sample. Hence, we propose a loss on text prompt embeddings to minimize intra-class text feature dispersion, while maximizing inter-class dispersion. The idea is inspired from contrastive learning \cite{khosla2020supervised} in supervised training where the intra-class distance w.r.t. anchor is minimized and inter-class distance w.r.t. negative sample is maximized. The proposed loss can be combined with other prompt tuning methods for \eg PromptAlign \cite{abdul2024align}, DiffTPT \cite{feng2023diverse}, TDA \cite{karmanov2024efficient}, BoostAdapter\cite{zhang2024boostadapter} could integrate \tca for prompt calibration.. In supplementary, we show gains in accuracy and \ece when we incorporate \tca on top of PromptAlign \cite{abdul2024align} and DiffTPT \cite{feng2023diverse}.
\item 
\textbf{Superior Performance:} We perform extensive experiments across various datasets and \myclip based models, incorporating our proposed attributes aware prompt initialization, and proposed loss. We report an average performance on $11$ benchmark datasets improving the model calibration by $7.5\%$ over the baseline \tpt\cite{tpt} and $2.01\%$ in terms of \ece over \ctpt \cite{yoon2024ctpt} respectively. 

\end{enumerate}

\section{Related Works}

\myfirstpara{Miscalibration in Neural Network}
Accurate estimation of predictive uncertainty, often referred to as model calibration, is a critical aspect of deploying neural networks in safety-sensitive applications. Proper calibration ensures that the confidence associated with a model's predictions aligns with its true accuracy, thereby facilitating more reliable decision-making. However, recent studies have highlighted frequent instances of miscalibration in modern neural network architectures, indicating a concerning trend: despite improvements in predictive performance, newer and more accurate models tend to produce poorly calibrated probability estimates \cite{calibrationmodern,wang2021rethinking}. 

\mypara{Calibration Techniques}
Calibration techniques can be broadly classified as train-time methods and post-hoc methods. Train-time techniques typically used additional loss terms along with the NLL (cross-entropy) loss during training. Some representative works include: \cite{hebbalaguppe2024calibration,pereyra2017regularizing,park2023acls,Patra_2023_WACV,hebbalaguppe2022novel,Hebbalaguppe_2022_CVPR,rawat2021pnpood,relcal}. These techniques are not practical in our setting as it requires retraining the neural network with the regularization terms.
Post-hoc calibration are applied after the model has been trained and often require  a validation set to fine-tune the output probabilities. Some common post-hoc calibration techniques include: \texttt{TS} \cite{platt}, \texttt{DC} \cite{kull2019beyond} etc. 

\mypara{Prompt Tuning for \vlms}
%
To efficiently adapt the large foundational models, prompting \citep{pt1} has emerged as a resource-efficient method. Prompt tuning typically uses static or learnt prompts as part of the input text to guide the model in performing specific tasks in a zero-shot, or few-shot manner. 
%
%
Hand-crafted prompts consisting of predefined vocabulary tokens, or hard prompts, may not be optimal in various settings. Hence, there is a growing focus on techniques that regard prompts as learnable vectors which can be optimized through gradient descent \cite{pt3}. 
For instance, \texttt{CoOp} \citep{coop} tunes the prompts in \myclip using labeled training samples to improve its classification accuracy. However, \texttt{CoCoOp} \citep{cocoop} identified that \texttt{CoOp} struggles with generalizing to out-of-distribution data and recommends conditioning the prompt on input images. While effective, these methods require access to annotated training data, which limits the zero-shot adaptation of pre-trained models like ours. To tackle this challenge, recent research has introduced a \tpt technique \citep{tpt}, which enables adaptive prompt learning at the inference time, using just one test sample. \tpt optimizes the prompt by minimizing the entropy with confidence selection so that the model has consistent predictions for each test sample. \texttt{DiffTPT} \cite{feng2023diverse} innovates test-time prompt tuning by leveraging pre-trained diffusion models to augment the diversity of test data samples used in \tpt. \texttt{PromptAlign} \cite{abdul2024align} fine-tunes multi-modal prompts at test-time by aligning the distribution statistics obtained from multiple augmented views of a single test image with the training data distribution statistics. Although previous studies \cite{coop,cocoop,plot,tpt} have primarily concentrated on refining prompt templates to improve accuracy, they have largely neglected calibration \cite{calibrationmodern}, except for \cite{yoon2024ctpt}. 

Our paper focuses on the critical and under-explored challenge of calibrating \vlms in a \textbf{zero-short, test-time setting}. To maintain the efficiency and practicality, we develop our solution within prompt tuning framework.

\section{Proposed Method}

\subsection{Preliminaries}

\myfirstpara{Confidence Calibration} 
Given a data distribution $\cD$ of $(x,y) \in \cX \times \{0,1\}$, let $c$ denote the predictive confidence of a predictor $f: \cX \rightarrow [0,1]$. The predictor  is said to be calibrated \cite{dawid}, if:
\begin{equation}
\bE_{(x,y) \sim \cD} \big[ y \mid f(x)= c \big] = c, \quad \forall c \in [0,1].    
\end{equation}
Intuitively, if a network predicts a class ``cancer'' for an image with a score of $0.9$, then a network is calibrated, if the probability that the image actually contains a cancer is $0.9$.
Expected Calibration Error (\ece) is a common metric used for measuring calibration, and evaluates how well the predicted confidence of a model align with its accuracy. To compute \ece, the confidence interval $[0, 1]$ is divided into a fixed number of bins. Each bin encompasses a range of predicted confidence. \ece value is then computed as \cite{Naeini}:
\begin{equation}
	\label{eq:ece}
    \text{ECE} = \sum_{k=1}^K \frac{|B_k|}{m} \left|\text{acc}(B_k) - \text{conf}(B_k)\right|,
    \nonumber
\end{equation}

where $K$ is the number of bins, $B_k$ is the set of samples, $|B_k|$ is the number of samples, $\text{acc}(B_k)$ is the prediction accuracy, and $\text{conf}(B_k)$ is the average predictive confidence in bin $k$. A lower \ece is preferred.


\mypara{Zero-Shot Classification with \myclip}
Let $\cX$ be the image space, and $\cY$ be the label space. Let $t \in T$ be the text prompt corresponding to an image sample $x \in \cX$. \myclip~\cite{clip} architecture is composed of two distinct encoders: a visual encoder denoted by: $f$, and a text encoder $g$. In the vanilla zero-shot inference with \myclip, we attach a manually designed prompt prefix, $\pb$ (e.g., $\pb =$ ``a photo of a") to each possible class $y_i \in \cY = \{y_1, y_2, \dots, y_K\}$, generating class-specific textual descriptions $t_i = [\pb; y_i]$. Here, $K$ denotes the number of classes. Next, we generate text features $g(t_i)$, and image features $f(x)$ by passing the relevant inputs to the respective encoders. This allows to compute the similarity between text feature, and image features as: $s_i = s \left( f(x), g(t_i=[\pb; y_i]\right)$, where $s(\cdot)$ refers to the cosine similarity. The probability of predicting class $y_i$ for the test image $x$ can be computed as: 
\[
p(y_i | x) = 
\frac{\exp\left( s \big( g(t_i),f(x) \big)/\tau \right)}
{\sum_{j=1}^K \exp\left( s \big( g(t_j),f(x) \big)/\tau \right)},
\] 

where $\tau$ is the temperature for the \texttt{softmax} function. The predicted class is $\yh = \argmax_{y_i} p(y_i \mid x)$, with predicted confidence $\ph = \max_{y_i} p(y_i \mid x)$.

\begin{figure}[ht]
\centering
\includegraphics[width=1\linewidth, height=4cm]{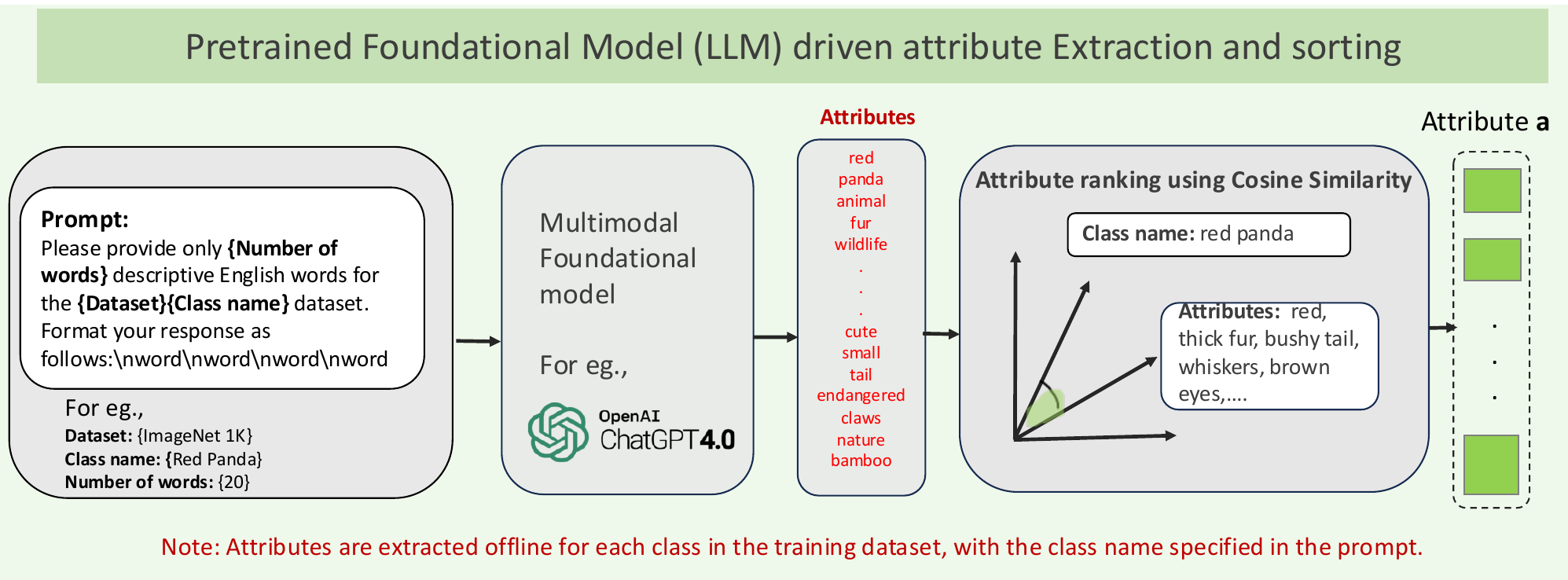}
  \caption{Visual attributes are extracted by prompting a multimodal foundational model as shown in the leftmost block. The extracted attributes (shown in red) are ranked based on their similarity to the Class name in the Dataset (e.g., the top 20 attributes for "red panda" in ImageNet1K dataset). This offline process aids model calibration by identifying relevant attributes. The relevant attributes $a \subset \{a_i\}_{i=1}^N$ by identifying the attribute similarity with respect to a class name. Here $a_i$ is the set of attributes returned for a particular class by pretrained \llm.}
 \label{fig:attExtraction}
\end{figure}

\mypara{Test-time Prompt Tuning} 
Several researchers have demonstrated the efficacy of few shot prompt tuning in general \cite{prompt_tuning_power,visual_prompt_tuning,intapt,neutral_prompttuning,wavelet}, as well as for \myclip based models \cite{coop,cocoop,plot,kgcoop}. 
%
Test-time prompt tuning (Vanilla \tpt (\vtpt)) introduced by \citep{tpt} aims to benefit from the rich knowledge of \myclip to boost its generalization in a zero-shot manner.  optimizes prompts without requiring labeled data. During inference, $N$ augmented views, $x^j$, of the test sample $x$ are generated. Predictions with entropy values below a predefined threshold are retained, while those with higher entropy are discarded through a confidence selection filter. The entropy of the remaining predictions is then averaged, and this value is used to update the prompts in an unsupervised manner using back-propagation from the following the objective function \cite{tpt}.
%
\begin{equation}
\cL_{\text{TPT}} = - \sum_{i=1}^{K} \pbar(y_i) \log \pbar(y_i), \quad \text{where} ~ \pbar(y_i) = \frac{1}{N} \sum_{j=1}^{N} p(y_i \mid x^j).
\end{equation}

Here, $\pbar(\cdot)$ represents the mean of vector class probabilities produced by the model across different augmented views preserved after the confidence selection filter. Additionally, it has been shown that test-time prompt tuning can be effectively combined with few-shot prompt tuning techniques (during train time), further boosting vanilla \vtpt's performance \citep{tpt}.

\begin{figure}[t!]
\centering
\includegraphics[width=\linewidth, height=6cm]{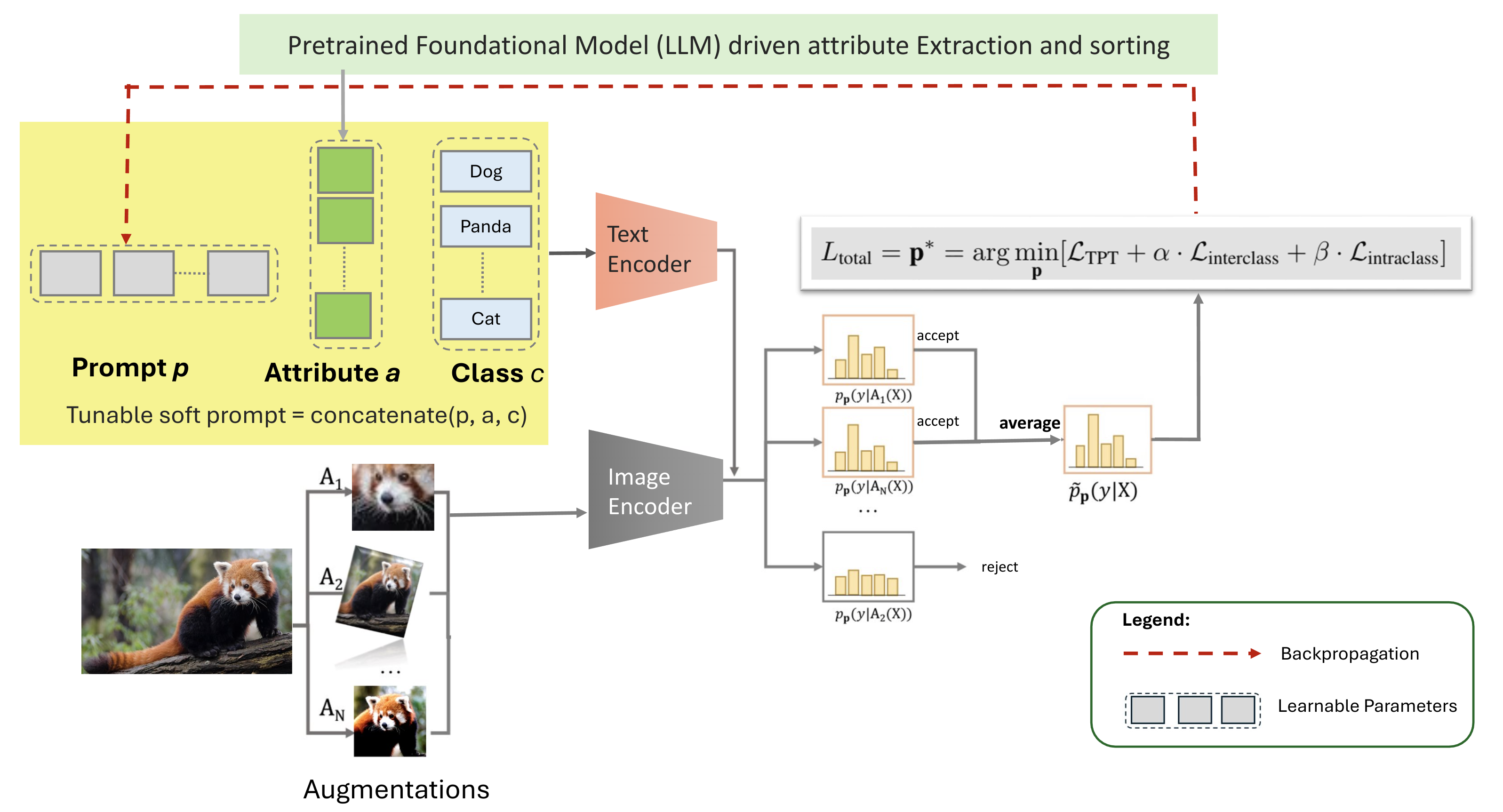}
 \caption{\textbf{Calibration using Test-time  Attribute Alignment for zero-shot image classification:} In a typical test time prompt tuning for image classification, a category label is prefixed with a template text, such as “a photo of a” (e.g., “a photo of a red panda”) to generate the prompt for tuning. Our approach differs in the following ways: \textbf{(a)} Visual attributes are extracted as shown in Fig. \ref{fig:attExtraction}. \textbf{(b)} Our approach takes an image and its augmentations ($A_1, A_2, ... A_N$) as the input. In contrast to \tpt\cite{tpt}, we utilize the attribute vector \textbf{a} concatenated with template text \textbf{p} and class name $c_i$ to initialize the prompt. We introduce two auxiliary terms in the objective function for test-time calibration via attribute alignment: $L_\text{interclass}$ to maximize mean text features between classes and $L_\text{intraclass}$ to minimize intra-class variance of textual attributes during prompt tuning to improve alignment between predicted and actual class probabilities, enhancing model calibration. This allows us to tune adaptive prompts on the fly with a single test sample, and without the need for additional training data or annotations. Both visual and text encoders are kept frozen while prompt tuning.}
 \label{fig:proposedMethod}
\end{figure}

\mypara{Attribute Alignment using an \llmtt} In \vlms, attribute alignment in prompt tuning guides the model to generate outputs matching specific visual or textual attributes. Authors in \cite{pratt2023does} use \llms to create descriptive sentences highlighting key features of image categories. An attribute extractor identifies relevant domain-specific information like color or context \cite{pratt2023does,nilsback2008automated,tian2024argue}, and the prompt is adjusted accordingly. This aligned prompt improves inference accuracy by tailoring the model to the task. Unlike the train-time techniques above, our approach focuses on test-time calibration.
%


\subsection{Test Time Calibration via Attribute Alignment}


Our proposed attribute-aware prompt tuning procedure comprises of two steps, namely, \textbf{(a)} relevant attribute extraction (See Fig \ref{fig:attExtraction}) ; \textbf{(b)} enhancing calibration via test-time loss on textual features separation/contraction (See Fig \ref{fig:proposedMethod}).

\cref{fig:attExtraction} depicts the first step, we obtain visual attributes that provide context by prompting \llms with inquiries about the visual characteristics of specific classes. The \llmtt input exclusively consists of class names from a dataset. Formally, given any label $y_i \in \cY$, we retrieve its corresponding class name, $c_i$, and a list of attributes $\textbf{a}_{y_i} = \gamma(y_i)$ where $\gamma$ is any language model like \gpt. The template for prompting \llmtt has been pre-defined (see Fig. \ref{fig:attExtraction}).  The attributes are subsequently ranked in descending order of relevance by sorting based on the cosine similarity between the class name and attribute names. We then store $M$ most relevant attributes in the attribute vector $\textbf{a}_c$ (we use top 2 attribute in our implementation based on our ablation study). In \cref{fig:attExtraction} we illustrate this with an example of a ``red panda" image. The attributes thus generated are appended to the tunable prompt, \pb\footnote{recall \pb is generated from manual template text, such as ``a photo of"}, along with the class names, such that tunable prompt = concatenate(\pb, \textbf{a}, c)) (also see the block diagram corresponding to yellow box in \cref{fig:proposedMethod}. The full prompt text including the attributes are shown in the \texttt{json} file for \texttt{Caltech 101} dataset included in the supplementary material. In step \textbf{(b)}, to enforce effective calibration, we employ a contrastive loss at test-time, and a test-time calibration process as specified in \cref{alg:conTextDispersion}. 

We start with the initialized prompts as described earlier, and then for every class $i$ and attribute $j$, we form the text embedding $\pb \oplus a_j \oplus c_i$ and then compute the centroid of these text embeddings. We then minimize the distance between class centroid and textual embeddings corresponding to class (generated using different class attributes). This is referred to as intra-class loss and serves to learn most discriminative features of a class. Similar to \ctpt \cite{yoon2024ctpt}, we also increase the distance between text embeddings of distinct classes and this loss is referred to as inter-class loss. For this, we first take the mean of the embeddings corresponding to different attributes of a specific class. This represents the textual feature corresponding to a class. We then maximise the distance between these representative features of each class so that all classes are well separated. The overall loss used to tune the prompts is the summation of vanilla test time prompt tuning loss $\cL_\tpt$\cite{tpt}, and the above two loss terms. Note that the back-propagated gradients only update tokens corresponding to $\pb$, whereas $\textbf{a}$, and $c_i$ tokens remain frozen, to prevent overfitting on the test sample. 

\begin{algorithm}[t]
\small
\caption{\textbf{T}est-time \textbf{C}alibration via \textbf{A}ttribute Alignment (\textbf{Inference})}
\begin{algorithmic}[1]
\State Initialize manual prompt, $\pb = $``a photo of a"
\State Attribute $a$ and class = $c$
\For{each class $i \in \{1,\ldots,K\}$}
	\For{each attribute $j \in \{1,\ldots,M\}$}
	    \State Form text embedding $t_{ij} = \pb \oplus \textbf{a}_{j} \oplus c_i$
	\EndFor
	\State Compute the mean of text embeddings for each class $\tbar_{y_i} = \frac{1}{M} \sum_{j=1}^{M} g(t_{ij})$, where $g(\cdot)$ is the \myclip text encoder.
	\State Calculate mean text attribute spread (\mtas) for class $y_{i}$: $\mtas(y_{i}) = \frac{1}{M} \sum_{j=1}^{M} \norm{g(t_{ij}) - \tbar_{y_i}}_2$
	\State $\cL_\text{intra-class}(y_i) = \mtas(y_{i})$  
\EndFor
\State Compute the mean of text embeddings for all classes, $\tdbar = \frac{1}{K} \sum_{i=1}^{K} \tbar_{y_i}$
\State Calculate Average Text Feature Dispersion (\atfd) \cite{yoon2024ctpt} across all classes: $\atfd = \frac{1}{K} \sum_{i=1}^{K} \norm{\tdbar - \tbar_{y_i}}_2$.
\State  $\cL_\text{inter-class}$ = - \atfd 
\State $\cL_\text{total} = \cL_\tpt + \alpha.\cL_\text{inter-class} + \beta. \cL_\text{intra-class}$.
\end{algorithmic}
\label{alg:conTextDispersion}
\end{algorithm}

\subsection{Understanding the Role of \textbf{\tca} in Enhancing Calibration}


\tca improves representation quality by leveraging contrastive learning principles thus enabling the generation of high-quality, meaningful, and discriminative embeddings that effectively capture semantic similarity. This is achieved through a contrastive test-time loss with inter-class ($\cL_\text{inter-class}$) and intra-class ($\cL_\text{intra-class}$) loss terms. The model classifies new samples by aligning them with the closest class embeddings while simultaneously distinguishing them from other classes. We believe this alignment enhances calibration during test-time.

Specifically, recall that calibration aims to align predictive probabilities with the true likelihood of an event. \tca addresses this by aligning similar representations while simultaneously mitigating overconfidence, a key factor contributing to miscalibration. The use of the term (See \cref{alg:conTextDispersion} line 12) plays a critical role in this process by explicitly penalizing embedding overlap for dissimilar classes. This discourages the model from assigning overly confident probabilities to incorrect predictions, ensuring that extreme predictive probabilities (close to $0$ or $1$) are only assigned when the different classes are well-separated. (See \cref{alg:conTextDispersion} lines 8 and 9) takes care of aligning similar textual embeddings.

\subsection{Difference between \tca and other contemporary techniques} 
Although prompt tuning through \ctpt \cite{yoon2024ctpt} introduces enhancements in calibration, it still falls short in capturing nuanced class specific features which are important to disambiguate between classes, and thus necessary for uncertainty calibration. Though the sample specific labels are absent in the test time setting as ours, however we make a observation, and note that even then class specific information is indeed available. We make use of \llms to generate class attributes and then use the proposed technique to choose most representative attributes. In another big difference, we choose not to update these attribute features. In \ctpt, firstly the text prompt initialization is same for all the classes, and then all of them get updated updated by the test-time loss, leading to overfitting on the sample, and less than ideal calibration. In our case, the frozen attribute based features provide adequate grounding and prevent overfitting, whereas other learnable prompts allow to adapt to the particular sample, thus leading to better calibration through proposed \tca over the current state-of-the-art, \ctpt. Our approach also differs from that of \tpt \cite{tpt}, as they do not incorporate attribute auxiliary information from \llms, nor do they explicitly optimize for calibration. As a result, their method exhibits sub-optimal calibration performance.

\section{Experiments}
\label{experiments}

\begin{table*}[t]
\centering
\setlength{\abovecaptionskip}{0.1cm}
\setlength{\belowcaptionskip}{0.1cm}
\renewcommand\arraystretch{1.2}
\setlength\tabcolsep{8pt}
\caption{\textbf{Fine-Grained Classification}. Results for CLIP-RN50 and CLIP-ViT-B/16 are reported, providing the \textbf{Accuracy ($\uparrow$)} and \textbf{ECE ($\downarrow$)} metrics for different experimental configuration (please see main test for configuration details). The values highlighted in \textbf{bold} indicate the lowest \ece achieved following test-time prompt tuning and \textbf{underline} is the second best. \textbf{Note:} The full table, which includes comparisons with other contemporary methods, can be found in the supplementary material due to space limitations in the main paper - we ablate \tca loss with promptAlign\cite{abdul2024align} and DiffTPT\cite{feng2023diverse} to show gains on top of contemporary methods \texttt{PromptAlign} (NeurIPS'24) and \texttt{DiffTPT} (ICCV'23)}
\label{tab:main_result}
\resizebox{\textwidth}{!}{%
\begin{tabular}{l|l|ccccccccccc|c}
\hline
\textbf{Method} & \textbf{Metric} & \textbf{ImageNet} & \textbf{Caltech} & \textbf{Pets} & \textbf{Cars} & \textbf{Flower} & \textbf{Food101} & \textbf{Aircraft} & \textbf{SUN397} & \textbf{DTD} & \textbf{EuroSAT} & \textbf{UCF101} & \textbf{Average} \\ \hline\hline
\rowcolor[HTML]{FDFCEB} 
\cellcolor[HTML]{FDFCEB} & Acc. & 58.1 & 85.8 & 83.8 & 55.7 & 61 & 74 & 15.6 & 58.6 & 40 & 23.7 & 58.4 & 55.9 \\
\rowcolor[HTML]{FDFCEB} 
\multirow{-2}{*}{\cellcolor[HTML]{FDFCEB}CLIP-RN50\textsubscript{HardPrompt}} & ECE & 3.83
& 4.33 & 5.91 & 4.7 & \textbf{3.19} & 3.11 & \underline{6.45} &\underline{3.54} & \underline{9.91} & 15.4 & \textbf{3.05} & \underline{5.61} \\
\rowcolor[HTML]{FFF8F2} 
\cellcolor[HTML]{FFF8F2} & Acc. & 60.7 & 87 & 84.5 & 58 & 62.5 & 74.9 & 17 & 61.1 & 41.5 & 28.3 & 59.5 & 57.7 \\
\rowcolor[HTML]{FFF8F2} 
\multirow{-2}{*}{\cellcolor[HTML]{FFF8F2}+TPT\textsubscript{HardPrompt}} & ECE & 11.4 & 5.04 & \underline{3.65} & 3.76 & 13.4 & 5.25 & 16.1 & 9.24 & 25.7 & 22.5 & 12.4 & 11.7 \\
\rowcolor[HTML]{FBEEEF} 
\cellcolor[HTML]{FBEEEF} & Acc. & 60.2 & 86.9 & 84.1 & 56.5 & 65.2 & 74.7 & 17 & 61 & 42.2 & 27.8 & 59.7 & 57.8 \\
\rowcolor[HTML]{FBEEEF} 
\multirow{-2}{*}{\cellcolor[HTML]{FBEEEF}+TPT\textsubscript{HardPrompt}+C-TPT} & ECE & \underline{3.01} & \underline{2.07} & \textbf{2.77} & \textbf{1.94} & 4.14 & {\textbf{1.86}} & 10.7 & \textbf{2.93} & 19.8 & \underline{15.1} & {\underline{3.83}} & 6.2 \\
\rowcolor[HTML]{FFEFFC} 
\cellcolor[HTML]{FFEFFC} & Acc. & 58.72 &	86.69 & 86.21 & 55.95 &	64.47 &	75.38 &	17.04 & 60.02 &	39.59 &	31.32 &	61.04 & 57.85\\
\rowcolor[HTML]{FFEFFC} 
\multirow{-2}{*}{\cellcolor[HTML]{FFEFFC}\textbf{+TPT\textsubscript{HardPrompt}+TCA (2 Attribute)}} & ECE 
& \textbf{1.76}	& \textbf{1.79}	& 5.43	& \underline{3.35}	& \underline{3.7} & 	\underline{2.45} & 	\textbf{4.48}	& 4.32	& \textbf{8.16}	& \textbf{5.5} & 	4.33 & \textbf{04.11}\\
\hline

\rowcolor[HTML]{FFF8F2} 
\cellcolor[HTML]{FFF8F2} & Acc. & 61.1 & 87.4 & 83.2 & 59.2 & 61.4 & 76.2 & 17.9 & 62 & 42.8 & 28.4 & 60.2 & 58.2 \\
\rowcolor[HTML]{FFF8F2} 
\multirow{-2}{*}{\cellcolor[HTML]{FFF8F2}+TPT\textsubscript{Ensemble}} & ECE & 11.2 & 4.29 & 4.79 & 3.08 & 14.1 & 5.27 & 14.6 & 7.68 & 22.2 & 18.9 & 11.1 & 10.7 \\
\rowcolor[HTML]{FBEEEF} 
\cellcolor[HTML]{FBEEEF} & Acc. & 61.2 & 87.4 & 84 & 57.3 & 65.3 & 76 & 17.5 & 62.1 & 43.1 & 29.4 & 60.7 & 58.5 \\
\rowcolor[HTML]{FBEEEF} 
\multirow{-2}{*}{\cellcolor[HTML]{FBEEEF}+TPT\textsubscript{Ensemble}+C-TPT} & ECE & \underline{4.13} & \textbf{2.15} & \textbf{2.71} & \textbf{1.68} & \underline{3.6} & {\textbf{1.47}} & \underline{10.9} & \textbf{2.96} & \underline{15.7} & \textbf{8.7} & {\underline{3.27}} & 5.2 \\


\rowcolor[HTML]{FFEFFC} 
\cellcolor[HTML]{FFEFFC} & Acc. & 68.1	& 93.26 &	90.13 &	65.94 &	68.9 &	84.23 &	25.38 &	65.84 &	43.91 &	47.17 &	67.72 & \textbf{65.50} \\
\rowcolor[HTML]{FFEFFC} 
\multirow{-2}{*}{\cellcolor[HTML]{FFEFFC}\textbf{+TPT\textsubscript{Ensemble}+TCA (2 Attributes)}} & ECE & \textbf{1.88}	& \underline{3.09} & \underline{4.38} & \underline{3.93} & \textbf{3.57}	& \underline{1.91} &	\textbf{3.36} &	\underline{6.02} & \textbf{4.36} &	\underline{9.36} &	\textbf{2.71} & \textbf{4.05} \\


\bottomrule
\rowcolor[HTML]{FDFCEB} 
\cellcolor[HTML]{FDFCEB} & Acc. & 66.7 & 92.9 & 88 & 65.3 & 67.3 & 83.6 & 23.9 & 62.5 & 44.3 & 41.3 & 65 & 63.7 \\
\rowcolor[HTML]{FDFCEB} 
\multirow{-2}{*}{\cellcolor[HTML]{FDFCEB}CLIP-ViT-B/16\textsubscript{HardPrompt}} & ECE & 2.12 & 5.5 & \underline{4.37} & \underline{4.25} & 3 & 2.39 & 5.11 & 2.53 & 8.5 & 7.4 & 3.59 & 4.43 \\
\rowcolor[HTML]{FFF8F2} 
\cellcolor[HTML]{FFF8F2} & Acc. & 69 & 93.8 & 87.1 & 66.3 & 69 & 84.7 & 23.4 & 65.5 & 46.7 & 42.4 & 67.3 & 65 \\
\rowcolor[HTML]{FFF8F2} 
\multirow{-2}{*}{\cellcolor[HTML]{FFF8F2}+TPT\textsubscript{HardPrompt}} & ECE & 10.6 & 4.51 & 5.77 & 5.16 & 13.5 & 3.98 & 16.8 & 11.3 & 21.2 & 21.5 & 13 & 11.6 \\
\rowcolor[HTML]{FBEEEF} 
\cellcolor[HTML]{FBEEEF} & Acc. & 68.5 & 93.6 & 88.2 & 65.8 & 69.8 & 83.7 & 24 & 64.8 & 46 & 43.2 & 65.7 & \textbf{64.8} \\
\rowcolor[HTML]{FBEEEF} 
\multirow{-2}{*}{\cellcolor[HTML]{FBEEEF}+TPT\textsubscript{HardPrompt}+C-TPT} & ECE & 3.15 & {\underline{4.24}} & \textbf{1.9} & \textbf{1.59} & 5.04 & \textbf{3.43} & \underline{4.36} & {\textbf{5.04}} & 11.9 & 13.2 & {\textbf{2.54}} & 5.13 \\

\rowcolor[HTML]{FFEFFC} 
\cellcolor[HTML]{FFEFFC} & Acc. & 67.37 & 92.86	& 90.51	& 65.92	& 69.18 &	69.18 &	25.32 &	65.5 &	44.73 &	45.58 &	66.9 & \underline{63.91} \\
\rowcolor[HTML]{FFEFFC} 
\multirow{-2}{*}{\cellcolor[HTML]{FFEFFC}\textbf{+TPT\textsubscript{HardPrompt}+TCA (2 Attribute)}} & ECE & \textbf{2.27}	& \textbf{3.01}	& 6.3 &	7.85 &	\textbf{3.67} &	5.28 &	\textbf{3.6} &	7.17 &	\textbf{5.48} & \textbf{8.37}	& \underline{2.82} & \textbf{5.07}\\
\hline
\rowcolor[HTML]{FDFCEB} 
\cellcolor[HTML]{FDFCEB} & Acc. & 68.2 & 93.4 & 86.3 & 65.4 & 65.7 & 85.2 & V23.5 & 64 & 45.6 & 43 & 66.1 & 64.2 \\
\rowcolor[HTML]{FDFCEB} 
\multirow{-2}{*}{\cellcolor[HTML]{FDFCEB}CLIP-ViT-B/16\textsubscript{Ensemble}} & ECE & 3.7 & 6.16 & 4.88 & 7.09 & 6.01 & 3.78 & 4.56 & 4.01 & 13.8 & 6.01 & 4.05 & 5.82 \\
\rowcolor[HTML]{FFF8F2} 
\cellcolor[HTML]{FFF8F2} & Acc. & 69.6 & 94.1 & 86.1 & 67.1 & 67.6 & 85.1 & 24.4 & 66.5 & 47.2 & 44 & 68.5 & 65.5 \\
\rowcolor[HTML]{FFF8F2} 
\multirow{-2}{*}{\cellcolor[HTML]{FFF8F2}+TPT\textsubscript{Ensemble}} & ECE & 9.82 & 4.48 & 5.72 & 4 & 13.9 & 4.27 & 14.6 & 9.01 & 18.6 & 14.1 & 10.5 & 9.91 \\
\rowcolor[HTML]{FBEEEF} 
\cellcolor[HTML]{FBEEEF} & Acc. & 69.3 & 94.1 & 87.4 & 66.7 & 69.9 & 84.5 & 23.9 & 66 & 46.8 & 48.7 & 66.7 & 65.8 \\
\rowcolor[HTML]{FBEEEF} 
\multirow{-2}{*}{\cellcolor[HTML]{FBEEEF}+TPT\textsubscript{Ensemble}+C-TPT} & ECE & 4.48 & 3.14 & \textbf{1.54} & \textbf{1.84} & 5.77 & {\underline{2.38}} & 6.4 & \textbf{3.09} & 13.7 & \textbf{5.49} & {\underline{3.04}} & \underline{4.62} \\

\rowcolor[HTML]{FFEFFC} 
\cellcolor[HTML]{FFEFFC} & Acc. & 68.1 & 93.26 &	90.13 &	65.94 &	68.9 & 84.23	& 25.38	& 65.84	& 43.91	& 47.17	& 67.72 & \textbf{65.5}\\
\rowcolor[HTML]{FFEFFC} 
\multirow{-2}{*}{\cellcolor[HTML]{FFEFFC}\textbf{+TPT\textsubscript{Ensemble}+TCA 2 attributes}} & ECE & \textbf{1.88} &	\textbf{3.09} &	4.38 &	3.93 &	\textbf{3.57} &	\textbf{1.91} &	\textbf{3.36} &	6.02 &	\textbf{4.36} &	9.36 &	\textbf{2.71} & \textbf{4.05}\\
\bottomrule
\end{tabular}%
}
\end{table*}

This section outlines the benchmarks for assessing our method and the experimental results. Consistent with previous research on the prompt tuning of vision-language models \cite{coop,cocoop,plot,tpt}, our evaluation is centered on two primary aspects: \textbf{(1)} a range of fine-grained classifications and \textbf{(2)} the natural distribution shift. \textbf{\textcolor{blue}{Note:}} In particular, given our objective to enhance calibration in the context of test-time prompt tuning, our experimental framework emphasizes prompt optimization in the absence of labeled training data.

\mypara{Datasets} 
For fine-grained classification, we utilize a diverse set of datasets, including ImageNet \cite{deng2009imagenet}, Caltech101 \cite{fei2004learning}, OxfordPets \cite{Pets}, StanfordCars \cite{Cars}, Flowers102 \cite{nilsback2008automated}, Food101 \cite{food101}, FGVCAircraft \cite{Aircraft}, SUN397 \cite{SUN397}, UCF101 \cite{UCF101}, DTD \cite{DTD}, and EuroSAT \cite{helber2019eurosat}. For the out-of-distribution (\ood) generalization task, we define ImageNet \cite{deng2009imagenet} as the in-distribution (source) dataset and extend evaluation to four \ood variants: ImageNetV2 \cite{imagenetV2}, ImageNet-Sketch \cite{wang2019learning}, ImageNet-A\cite{imagenetA}, and ImageNet-R\cite{imagenetR}.

\mypara{Implementation Details}
We report results in following experimental configurations. The initialized prompt is set to a hard prompt ‘\texttt{a photo of a}’ (CLIP\tsub{HardPrompt}) and the corresponding 4 tokens are optimized based on a single test image using \tpt (TPT\tsub{HardPrompt}) or jointly using \tpt and our proposed technique \tca (TPT\tsub{HardPrompt})+TCA). We also include an ensemble setting where we average the logits from 4 different hard-prompt initialization using {‘\texttt{a photo of a}’, ‘\texttt{a photo of the}’, ‘\texttt{a picture of a}’, ‘\texttt{a picture of the}’} (CLIP\tsub{Ensemble}). Similarly, we optimize using \tpt as well (TPT\tsub{Ensemble}), or jointly using \tpt and \tca (TPT\tsub{Ensemble}+TCA) on each of the hard-prompt initialization and average the resulting logits. We have tried to use 1, 2, and 3 attribute initialization. 
\textbf{Hyperparameters $\alpha$ and $\beta$:} We employ a test-time prompt tuning strategy, which does not allow access to data for hyperparameter tuning. We perform a grid search over $\alpha$ and $\beta$  to balance the calibration loss for the least ECE using Caltech 101 dataset and apply the same values for 11 datasets following a setup similar to C-TPT\cite{yoon2024ctpt}. We obtain $(\alpha,\beta)$ as $(10, 35)$, respectively. Using 2 attributes gave the best ECE values on majority of the datasets for finegrained classification. For Natural distribution shifts, we obtained, $(\alpha,\beta)$ as $(45, 15)$. 
For \tpt \cite{tpt}, we optimize the prompt in one step using the AdamW optimizer with a learning rate of 0.005. Our method runs on a single NVIDIA Tesla V100 GPU with 32GB of memory, except for the ImageNet, ImageNet-A, and ImageNet V2 datasets, which use two GPUs for evaluation. 

\subsection{Comparison on Fine Grained Classification}

For the fine-grained classification task, we compare contemporary methods against hard prompt and benchmark approaches, such as \tpt \cite{tpt} and \ctpt \cite{yoon2024ctpt}. Tab. \ref{tab:main_result} summarizes the results: accuracy and \ece values. Our evaluation includes multiple \myclip architectures, specifically \myclip RN-50 and ViT-B/16. The results show that our method significantly outperforms the hard prompt configuration. 
When comparing the average performance of \ctpt across all $11$ datasets, our method achieves a similar average predictive accuracy while notably reducing the average \ece. For \myclip RN-50, the \ece decreases from $5.6$ to $4.11$. Similarly, for ViT-B/16, the \ece is reduced from $5.82$ to $4.05$.

\begin{table}[t]
\centering
\caption{\textbf{Natural Distribution Shifts}. Results for CLIP-RN50 and CLIP-ViT-B/16 are reported, providing the \textbf{Acc. ($\uparrow$)} and \textbf{ECE ($\downarrow$)} metrics for different experimental configurations (please refer to the main text for details of configurations). Dataset abbreviations: ImageNet-V2 (IN-V2), ImageNet-A (IN-A), ImageNet-R (IN-R), and ImageNet-Sketch (IN-S). Values highlighted in \textbf{bold} indicate the lowest \ece achieved after test-time prompt tuning. 
}
\label{table:naturalshift}
\small 
\setlength{\tabcolsep}{4pt} 
\resizebox{0.8\linewidth}{!}{%
\begin{tabular}{l|l|cccc|c}
\toprule
\textbf{Methods} & \textbf{Metric} & \textbf{IN-A} & \textbf{IN-V2} & \textbf{IN-R} & \textbf{IN-S} & \textbf{Avg.} \\ \hline \hline
\rowcolor[HTML]{FDFCEB} 
\cellcolor[HTML]{FDFCEB} & Acc. & 21.7 & 51.4 & 56 & 33.3 & 40.6 \\
\rowcolor[HTML]{FDFCEB} 
\multirow{-2}{*}{\cellcolor[HTML]{FDFCEB}CLIP-RN50\textsubscript{HardPrompt}} & ECE & 21.3 & 3.33 & 2.07 & 3.15 & 7.46 \\
\rowcolor[HTML]{FFF8F2} 
\cellcolor[HTML]{FFF8F2} & Acc. & 25.2 & 54.6 & 58.9 & 35.1 & 43.5 \\
\rowcolor[HTML]{FFF8F2} 
\multirow{-2}{*}{\cellcolor[HTML]{FFF8F2}+TPT\textsubscript{HardPrompt}} & ECE & 31.0 & 13.1 & 9.18 & 13.7 & 16.7 \\
\rowcolor[HTML]{FBEEEF} 
\cellcolor[HTML]{FBEEEF} & Acc. & 23.4 & 54.7 & 58 & 35.1 & 42.8 \\
\rowcolor[HTML]{FBEEEF} 
\multirow{-2}{*}{\cellcolor[HTML]{FBEEEF}+TPT\textsubscript{HardPrompt}+C-TPT} & ECE & 25.4 & 8.58 & 4.57 & 9.7 & 12.1 \\
\rowcolor[HTML]{FFEFFC} 
\cellcolor[HTML]{FFEFFC} & Acc. & 20.77	& 51.74 &	54.83 &	32.83 & 40.04  \\
\rowcolor[HTML]{FFEFFC} 
\multirow{-2}{*}{\cellcolor[HTML]{FFEFFC}\textbf{+TPT\textsubscript{HardPrompt}+TCA (2 Attributes)}} & ECE &	\textbf{22.53}	& \textbf{4.39} &	\textbf{1.25} &	\textbf{6.22} & \textbf{8.59} \\
\hline
\rowcolor[HTML]{FDFCEB} 
\cellcolor[HTML]{FDFCEB} & Acc. & 22.7 & 52.5 & 57.9 & 34.7 & 42 \\
\rowcolor[HTML]{FDFCEB} 
\multirow{-2}{*}{\cellcolor[HTML]{FDFCEB}CLIP-RN50\_Ensemble} & ECE & 17 & 2.68 & 5.64 & 10.9 & 9.06\\
\rowcolor[HTML]{FFF8F2} 
\cellcolor[HTML]{FFF8F2} & Acc. & 26.9 & 55 & 60.4 & 35.6 & 44.5 \\
\rowcolor[HTML]{FFF8F2} 
\multirow{-2}{*}{\cellcolor[HTML]{FFF8F2}+TPT\textsubscript{Ensemble}} & ECE & 29.1 & 12.7 & 7.5 & 14 & 15.8 \\
\rowcolor[HTML]{FBEEEF} 
\cellcolor[HTML]{FBEEEF} & Acc. & 25.6 & 54.8 & 59.7 & 35.7 & 44 \\
\rowcolor[HTML]{FBEEEF} 
\multirow{-2}{*}{\cellcolor[HTML]{FBEEEF}+TPT\textsubscript{Ensemble}+C-TPT} & ECE &  27 &  9.84 &  5.17 &  12.2 &  13.6 \\
\rowcolor[HTML]{FFEFFC} 
 \cellcolor[HTML]{FFEFFC} & Acc. & 21.12 &	51.8 &	55.57 & 33.11 & 40.4 \\
\rowcolor[HTML]{FFEFFC} 
 \multirow{-2}{*}{\cellcolor[HTML]{FFEFFC}\textbf{+TPT\textsubscript{Ensemble}+TCA (2 Attributes)}} & ECE & \textbf{22.99}	& \textbf{3.69} &	\textbf{0.94} &	\textbf{5.37} & \textbf{8.24}\\
\hline


\rowcolor[HTML]{FDFCEB} 
\cellcolor[HTML]{FDFCEB} & Acc. & 47.8 & 60.8 & 74 & 46.1 & 57.2 \\
\rowcolor[HTML]{FDFCEB} 
\multirow{-2}{*}{\cellcolor[HTML]{FDFCEB}CLIP-ViT-B/16\textsubscript{HardPrompt}} & ECE & 8.61 & 3.01 & 3.58 & 4.95 & 5.04 \\
\rowcolor[HTML]{FFF8F2} 
\cellcolor[HTML]{FFF8F2} & Acc. & 52.6 & 63 & 76.7 & 47.5 & 59.9 \\
\rowcolor[HTML]{FFF8F2} 
\multirow{-2}{*}{\cellcolor[HTML]{FFF8F2}+TPT\textsubscript{HardPrompt}} & ECE & 16.4 & 11.1 & 4.36 & 16.1 & 12 \\
\rowcolor[HTML]{FBEEEF} 
\cellcolor[HTML]{FBEEEF} & Acc. & 51.6 & 62.7 & 76 & 47.9 & 59.6 \\
\rowcolor[HTML]{FBEEEF} 
\multirow{-2}{*}{\cellcolor[HTML]{FBEEEF}+TPT\textsubscript{HardPrompt}+C-TPT} & ECE & \textbf{8.16} & 6.23 & \textbf{1.54} & \textbf{7.35} & \textbf{5.82} \\

\rowcolor[HTML]{FFEFFC} 
\cellcolor[HTML]{FFEFFC} & Acc. & 46.95 &	59.94 &	72.78 &	45.1 & 56.19\\
\rowcolor[HTML]{FFEFFC} 
\multirow{-2}{*}{\cellcolor[HTML]{FFEFFC}\textbf{+TPT\textsubscript{HardPrompt}+TCA (2 Attributes)}} & ECE & 8.59 & \textbf{4.95} &	5.1 &	8.62 & 6.81\\



\hline
\rowcolor[HTML]{FDFCEB} 
\cellcolor[HTML]{FDFCEB} & Acc. & 50.9 & 62 & 74.5 & 46 & 58.4 \\
\rowcolor[HTML]{FDFCEB} 
\multirow{-2}{*}{\cellcolor[HTML]{FDFCEB}CLIP-ViT-B/16\textsubscript{Ensemble}} & ECE & 8.85 & 3.01 & 2.85 & 9.7 & 6.1 \\
\rowcolor[HTML]{FFF8F2} 
\cellcolor[HTML]{FFF8F2} & Acc. & 54.2 & 63.9 & 78.2 & 48.5 & 61.2 \\
\rowcolor[HTML]{FFF8F2} 
\multirow{-2}{*}{\cellcolor[HTML]{FFF8F2}+TPT\textsubscript{Ensemble}} & ECE & 13.5 & 11.2 & 3.64 & 15.3 & 10.9 \\
\rowcolor[HTML]{FBEEEF} 
\cellcolor[HTML]{FBEEEF} & Acc. & 52.9 & 63.4 & 78 & 48.5 & 60.7 \\
\rowcolor[HTML]{FBEEEF} 
\multirow{-2}{*}{\cellcolor[HTML]{FBEEEF}+TPT\textsubscript{Ensemble}+C-TPT} & ECE & 10.9 & 8.38 & \textbf{1.4} & 12.6 & 8.32 \\

\rowcolor[HTML]{FFEFFC} 
\cellcolor[HTML]{FFEFFC} & Acc. & 47.36	& 60.85 &	72.74 & 45.72& 56.66 \\
\multirow{-2}{*}{\cellcolor[HTML]{FFEFFC}+TPT\textsubscript{Ensemble}+TCA (2 attributes)} & ECE & \textbf{5.21}	& \textbf{1.81} &	3.42 &	\textbf{4.81} & \textbf{3.81}\\ 
\hline
\end{tabular}
}
\end{table}

\subsection{Robustness to Natural Distribution Shifts}
\label{sec:exp-ood}

We follow the setting in Radford \etal \cite{clip} and evaluate model's robustness to natural distribution shifts on 4 ImageNet Variants which have been considered as \ood for ImageNet in previous works. We report the results in \cref{table:naturalshift}. The table shows that we outperform contemporary methods (\tpt, and \ctpt) in terms of \ece on 3 out of 4 datasets.


\subsection{Ablation Study} 

We investigate the factors contributing to calibration— whether it is driven by the inclusion of attributes or by the choice of loss function. To examine this, we conducted an experiment under two conditions. In the first condition, we incorporate attributes into the prompts and evaluate the method using the \tpt loss function. In the second, we again incorporate attributes into the prompts but evaluate using the combined \tpt+\tca loss function on $3$ datasets. 

\begin{figure}[t]
    \centering
    \includegraphics[width=\linewidth]{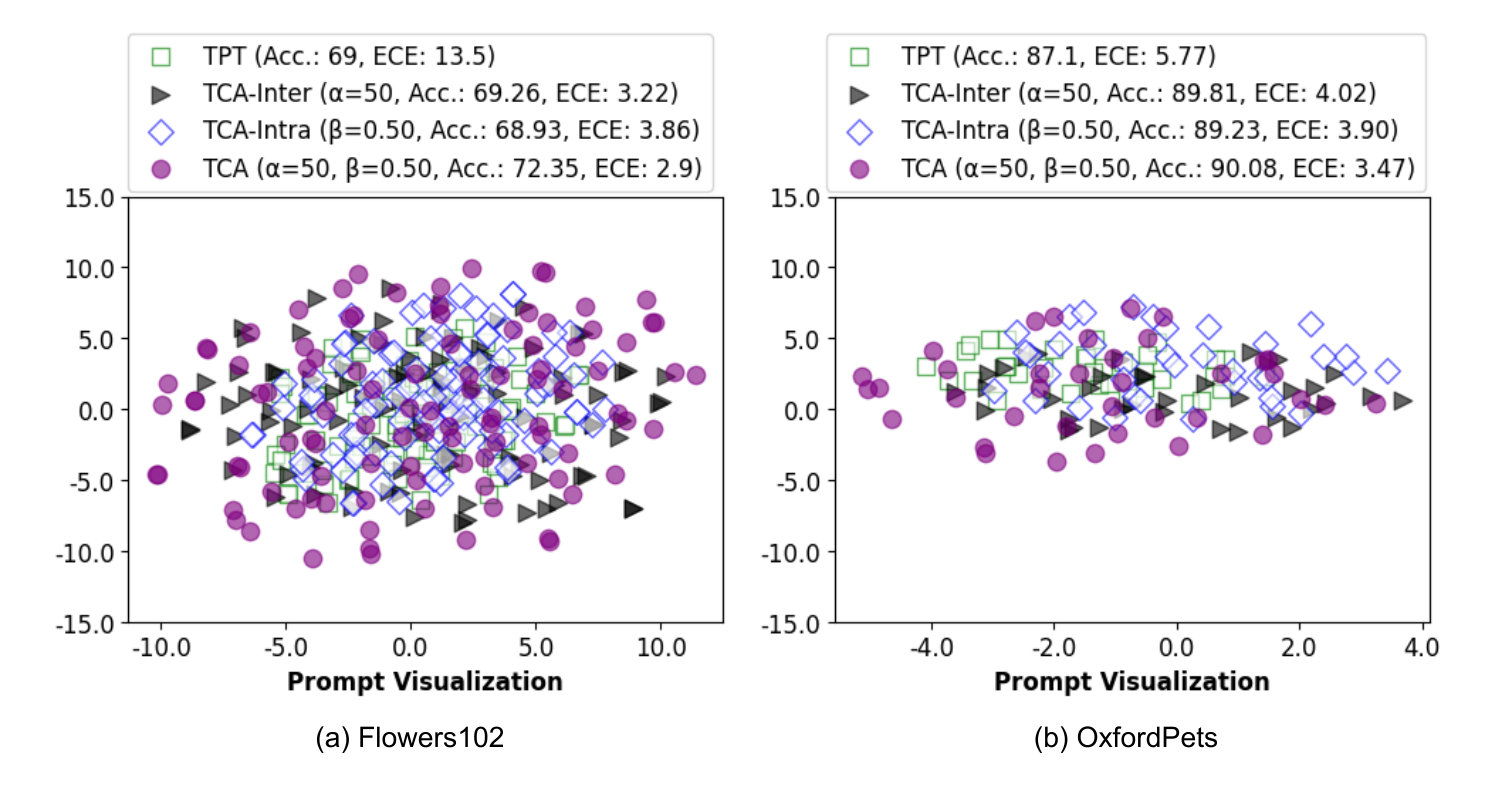}
    \caption{ The t-SNE plot shows Class-specific Text Embeddings  on tuned prompts. We conduct ablation on each term of $L_{\text{total}} = \textbf{p}^* = \arg \min_{\textbf{p}}[\cL_\tpt + \alpha.\cL_\text{inter-class} + \beta. \cL_\text{intra-class}]$ to understand its relative contribution empirically. In (a) and (b), notice that incorporating all three terms in $L_{\text{total}}$
  results in the lowest \ece and highest feature dispersion or spead.}
    \label{fig:dispersion}
\end{figure}

\mypara{Relative Contribution of Attribute Initialization and Proposed Loss}
To better understand the contribution we conduct the ablation experiments on \texttt{DTD} dataset using ResNet50 feature extractor and report (Acc $\uparrow$, \ece $\downarrow$). We have 3 variants: (a) $+\tpt_{HardPrompt}$ ($41.5,25.7$), (b) $+\tpt_{HardPrompt}$+ initialization with 2 attributes ($40.96,20.45$), (c) $+\tpt_{HardPrompt}$ + initialization with 2 attributes + proposed \tca loss ($42.79,5.59$). The key observations with ablation are as follows: \textbf{(1.)} \textbf{Attribute Initialization:} When initialized with 2 attributes, there was a $20.6\%$ reduction in \ece compared to the hard prompt model; \textbf{(2.) }\textbf{\tca Loss:} Introduction of the \tca loss resulted in a $3.65\times$ reduction in \ece, bringing \ece down from $25.7$ to $5.59$, significantly improving the model's calibration. \textbf{(3.)} \textbf{Combined Effect of Both:} When both attribute initialization and \tca loss were used together, the \ece reduction was even more pronounced, with an overall $4.59\times$ reduction in \ece, yielding the lowest \ece value of $5.59$ and maximum accuracy of $42.79$.
Thus, both proposed contributions, attribute initialization strategy, as well as the proposed loss play significant roles in improving model calibration. The proposed loss is particularly effective in reducing \ece, and combining it with attribute initialization leads to the most significant improvement in both accuracy and calibration. Refer to \cref{fig:dispersion}, which illustrates the comparison of feature dispersion, found to be inversely correlated with \ece. When both inter- and intra-loss terms are utilized, we observe the maximum Class-specific Text Embedding dispersion and the lowest \ece, consistent with the findings of \cite{yoon2024ctpt}. See suppl. for details on how the plot was obtained.


%


\subsection{Discussion}

\noindent \textbf{Confidence Calibration and \tca:} 
 Here, we provide an intuitive understanding of our proposed loss function, formulated as: $L_{\text{total}} = \textbf{p}^* = \arg \min_{\textbf{p}}[\cL_\tpt + \alpha.\cL_\text{inter-class} + \beta. \cL_\text{intra-class}]$. To assess the significance of each component within this formulation, we conduct a systematic ablation study. This includes t-SNE visualizations, which facilitate the analysis of the impact of individual loss terms on feature separability and clustering. Additionally, we compare our approach against state-of-the-art test-time calibration methods in the zero-shot setting, thereby demonstrating its effectiveness and robustness.

\noindent \textbf{Need for intra-inter class losses:} \tca improves representation quality by leveraging contrastive principles thus enabling the generation of high-quality, discriminative embeddings that effectively capture semantic similarity/dissimilarity.  \tca addresses calibration by aligning similar classes, and the use of the dispersion term explicitly penalizes the embedding overlap for dissimilar classes. This discourages the model from assigning overly confident probabilities to incorrect predictions, ensuring that extreme predictive probabilities (near $0$ or $1$) are only assigned when the different classes are well-separated. Fig. \ref{fig:dispersion} shows an ablation over individual loss terms' impact on calibration: Using both $\cL_{\text{intra}}$ and $\cL_{\text{inter}}$ in $\cL_{\text{total}}$ leads to the lowest ECE and greatest text feature dispersion.

\noindent \textbf{Conceptual differences between \tca Loss and Contemporaries:} The recent contemporary method, DAPT\cite{cho2023distribution} targets improved accuracy in few-shot settings, whereas we focus on zero-shot calibration. DAPT uses exponential inter- and intra-dispersion on both vision and text embeddings, while our method relies on $L_2$ norm distance between the test sample and mean text embeddings. $L_2$ norm is easier to interpret as it measures the Euclidean distance between embeddings, making it more intuitive and transparent, especially when comparing distances in high-dimensional spaces, but less sensitive to outliers and computationally efficient. \cite{levine2023enabling}  facilitates calibration using temperature scaling on the ImageNet validation set. However, when applying TS with TCA loss on the Caltech 101 dataset (ViT B-16), we observe a degradation in (Accuracy,ECE) from 93.02, 12.92 with TS vs. 92.45, 3.89 without TS, suggesting a decrease in performance with TS. \cite{wang2024open} uses Distribution aware calibration for fine-tuned VLM calibration, while our focus is on zero-shot settings like C-TPT\cite{yoon2024ctpt}. Finally, \cite{oh2024towards} involves few-shot finetuning, making it not directly comparable to our approach. 

\mypara{Vizualisation of Class-specific Text Embeddings  on tuned prompts}
Please refer to the supplemental materials for t-SNE plots across multiple datasets, which illustrate the lower \ece and the highest dispersion indicating better class separability of \tca relative to contemporaneous methods.

\noindent \textbf{Supplementary Material}
details the factors behind \tca's superior performance, datasets, metrics, feature extractor, experimental setup, hyperparameters, and t-SNE comparisons with PromptAlign \cite{abdul2024align}, DiffTPT \cite{feng2023diverse}.

\section{Conclusions and Future directions}
\label{sec:conclusions}



In this work, we introduced two key insights to enhance the effectiveness of test time prompt tuning. First, we demonstrated that attribute-aware prompting, wherein relevant visual attributes are appended to the prompts. This allows the model to better align its visual embeddings with discriminative features, resulting in improved predictive uncertainty handling and class-separation. Second, we proposed a regularization loss that encourages the model to minimize intra-class text feature dispersion while maximizing inter-class dispersion, inspired by contrastive learning principles. This ensures that the learned prompts do not overfit to individual samples, even when limited data is available.

This work opens up new possibilities for leveraging unsupervised attribute information to improve model performance in low-data or test-time settings, paving the way for more robust and adaptable models in real-world applications. In future, it would be interesting to study the effectiveness on other VLM architectures apart from CLIP such as Flamingo.


\vspace{-40em}
\bibliographystyle{splncs04}
\bibliography{main2}

\newpage
\section{Supplemental material}
\label{sec:supp_intro}

 To keep the main manuscript self-contained, we include the following details:

 \begin{itemize}
     \item \textbf{Test-Time Prompt Tuning:} We present a detailed description of our loss function and provide insights into its formulation. Additionally, we provide an intuitive explanation of how integrating this loss function has the potential to enhance calibration.
     \item \textbf{Datasets:} We provide a comprehensive description of the datasets utilized for fine-grained classification and natural distribution shift here (see Table 1 and Table 2 of the main text).
     \item \textbf{Reproducible Research:} To facilitate reproducible research, following acceptance, we will make the source code publicly available.

     
     \item \textbf{Additional results}: In this study, we present results from the application of the \texttt{PromptAlign} test-time prompt tuning technique \cite{abdul2024align} and \ctpt \cite{yoon2024ctpt} across 10 datasets, and we compare its performance with our proposed approach, \tca. Our findings demonstrate that integrating \tca with \texttt{PromptAlign}\cite{abdul2024align} leads to a reduction in calibration error and an improvement in accuracy. Additionally, we provide t-SNE visualizations to further investigate the distribution of text features, which complement the datasets discussed in the main text.
 \end{itemize}

\section{Test-Time Prompt Tuning}
\label{sec:tpt}

\subsection{Background}
\label{subsec:background}
Test time prompt tuning or \tpt in short adapts a pre-trained language model (LLM/VLM) to specific tasks or domains during inference, eliminating the need for retraining or full fine-tuning. It aims to enhance the model's performance on a given task by adjusting its input prompts, all without modifying the model’s core parameters. In our setting, we aim to learn adaptively the prompts on the fly with a single test sample \cite{tpt}.

\subsection{Why  is \textbf{\tpt} attractive?}
\label{subsec:attractive_tpt}

\tpt is particularly appealing due to its ability to operate on a single test sample without the need for large training datasets or the extensive computational resources typically required for training-time calibrators. Additionally, \tpt offers significant advantages in terms of efficiency, as it requires less time and computational effort to adapt a sample for generalization and calibration.

\subsection{Challenges in Contemporary \textbf{\tpt} Approaches}
\label{subsec:calibTPT}


Despite the key advantages of \tpt such as dynamic adaptation, improved robustness to distributional shifts, the resource efficiency, these methods often encounter challenges in calibration, particularly in dynamically adapting to the diverse textual feature distributions encountered in real-world data. This limitation restricts their ability to achieve effective prompt calibration.

Several methods illustrate these shortcomings. For instance, \texttt{ArgGue}\cite{tian2024argue} utilizes argument-guided prompt learning to refine task-specific tuning but does not explicitly address calibration concerns. Similarly, \texttt{DiffTPT}\cite{feng2023diverse} focuses on generating diverse image variants to improve task adaptability; however, it overlooks the specific optimization of calibration metrics. \texttt{PromptAlign}~\cite{abdul2024align} aligns prompts with semantic features to enhance task performance but explicitly does not account for calibration. To this end, our goal is to enhance calibration without much trade-off in accuracy.


\subsection{\textbf{\tca:} Insights on our Proposed Loss function for Calibration}
\label{subsec:TCA}

As mentioned in the main text, to enforce calibration, we apply contrastive loss on textual attributes. We first follow (a) attribute extraction and ranking mentioned in Fig 2(a) of the main text. Subsequently,  we follow Alg. 1 (in the main text) to induce the test-time calibration. 

Within a class, we enforce minimization of textual attribute distances with respect to the centroid and among different classes we maximize the distance of per class mean embeddings. We list the terms we introduce on top of $\mathcal{L}_{TPT}$ to enforce calibration here. Our loss is a combination of interclass attribute dispersion and intraclass attribute contraction-- we term our loss function called Test-Time Calibration via Attribute Alignment (\texttt{TCA}), which incorporates both inter- and intra-class terms to facilitate prompt learning. 

Let the total number of classes be $K$, and the total number of attributes be $M$. Seeking inspiration from contrastive training,  We compute the mean of encoded text embeddings for each class $y_i$ as follows:

\begin{equation}
\tbar_{y_i} = \frac{1}{M} \sum_{j=1}^{M} g(t_{ij})
\end{equation}

where $g(\cdot)$ is the \myclip text encoder, $i$ and $j$ index class and attributes respectively.

Subsequently, we calculate mean text attribute spread (\mtas) for class $y_{i}$: 
\begin{equation}
\mtas(y_{i}) = \frac{1}{M} \sum_{j=1}^{M} \norm{g(t_{ij}) - \tbar_{y_i}}_2
\end{equation}
\mtas is analogous to \atfd as defined in \cite{yoon2024ctpt}, however, \mtas also incorporates attribute information for prompt initialization differentiating it from \cite{yoon2024ctpt} \footnote{Note: Average Textual Feature Dispersion (ATFD) refers to a metric used to evaluate the spread or diversity of textual features across different instances in a given dataset. Specifically, it measures how varied or dispersed the features of textual data are when mapped into a feature space. The idea is that, in a high-quality representation space, the features corresponding to similar texts should be close together, and the features for dissimilar texts should be more distant. ATFD, in this case, helps to quantify how dispersed or clustered the features are on average.}
\begin{equation}
   \cL_\text{intra-class}(y_i) = \mtas(y_{i}) 
   \label{equ:intra}
\end{equation}

We impose inter-class distance by first computing the mean of text embeddings for each class. This approach ensures that the class representations are well-separated, promoting distinctiveness across classes. The process is formally described as follows:

\begin{equation}
    \tdbar = \frac{1}{K} \sum_{i=1}^{K} \tbar_{y_i}
\end{equation}

Now, we calculate Average Text Feature Dispersion (\atfd) \cite{yoon2024ctpt} across all classes as follows:

\begin{equation}
\atfd = \frac{1}{K}\sum_{i=1}^{K} \norm{\tdbar - \tbar_{y_i}}_2
\label{equ:atfd}
\end{equation}

Similar to contrastive training, we aim to maximize the distance between representations of different classes, as formulated below:

\begin{equation}
\cL_\text{inter-class} = -\atfd 
\end{equation}

\paragraph{Total Loss:} 

\noindent The total loss for test-time calibration for zero-shot classification can be formulated as:
\begin{equation}
L_{\text{total}} = \textbf{p}^* = \arg \min_{\textbf{p}}[\cL_\tpt + \alpha.\cL_\text{inter-class} + \beta. \cL_\text{intra-class}].
\label{equ:totalLoss}
\end{equation}

Here, $\textbf{p}^*$ is the optimal prompt achieved through backpropagation using stochastic gradient descent and is aimed to optimize calibration. The loss terms, $\cL_\text{intra-class}$ and $\cL_\text{inter-class}$ are used to enforce intra-class feature contraction and maximize intraclass text feature dispersion (\textbf{Note:} $\mathcal{L_{\text{inter-class}}}$ = - \atfd).  $\alpha$ and $\beta$ are the hyperparameters to control the relative importance with respect to \texttt{inter-class} and \texttt{intra-class} losses.

\subsection{Understanding the Role of \textbf{\tca} in Enhancing Calibration}


\tca improves representation quality by leveraging contrastive learning principles thus enabling the generation of high-quality, meaningful, and discriminative embeddings that effectively capture semantic similarity. This is achieved through a contrastive test-time loss with inter-class ($\cL_\text{inter-class}$) and intra-class ($\cL_\text{intra-class}$) loss terms. The model classifies new samples by aligning them with the closest class embeddings while simultaneously distinguishing them from other classes. We believe this alignment enhances calibration during test-time.

Specifically, recall that calibration aims to align predictive probabilities with the true likelihood of an event. \tca addresses this by aligning similar representations while simultaneously mitigating overconfidence, a key factor contributing to miscalibration. The use of the term (See \cref{equ:atfd}) plays a critical role in this process by explicitly penalizing embedding overlap for dissimilar classes. This discourages the model from assigning overly confident probabilities to incorrect predictions, ensuring that extreme predictive probabilities (close to $0$ or $1$) are only assigned when the different classes are well-separated. \cref{equ:intra} takes care of aligning similar textual embeddings.


\section{Datasets}
\subsection{Non-semantic/Natural Distribution Shifts}
\label{sec:exp-ood}
\textbf{Datasets.} In order to evaluate the robustness wrt distribution shifts that can occur naturally in real-world scenarios, we follow the setting proposed in Radford et al. \citep{clip,yoon2024ctpt} to evaluate the model's robustness to natural distribution shifts on 4 ImageNet variants. These have been considered as out-of-distribution (OOD) data for ImageNet~\citep{deng2009imagenet} in previous work.
\begin{itemize}

    \item \textbf{ImageNet-Sketch}~\citep{wang2019learning} is a dataset of black and white sketches, collected independently from the original ImageNet validation set. The dataset includes 50,000 images in total, covering 1,000 ImageNet categories.
    
    \item \textbf{ImageNet-R}~\citep{imagenetR} collects images of ImageNet categories but with artistic renditions. There are 30,000 images in total, including 200 ImageNet categories.

    \item \textbf{ImageNet-V2}~\citep{imagenetV2} is an independent test set containing natural images, collected from different source, including 10,000 images of 1,000 ImageNet categories.
    \item \textbf{ImageNet-A}~\citep{imagenetA} is a challenging test set of ``natural adversarial examples"  consisting of 7,500 images of 200 of ImageNet categories.

\end{itemize}

\subsection{Datasets for Finegrained Classification}
\label{appendix:dataset}
The fine-grained classification experimental setup comprises $11$ datasets following \cite{yoon2024ctpt,tpt}. As mentioned in \cite{yoon2024ctpt} we summarize the number of classes and test-set size for each dataset in \cref{appendix:table_dataset}.

\begin{table}[ht]
\setlength\tabcolsep{8pt}
\centering
\caption{\textbf{The detailed statistics of datasets used in the experiments:} The datasets highlighted in \texttt{lavender} color are designated for fine-grained classification, whereas those without highlight are intended for classification under natural distribution shifts to assess robustness.}
\begin{tabular}{lccc}
\hline 
\rowcolor[HTML]{FDFCEB} Dataset & \# Classes & Test set size \\
\hline 
\hline 
\rowcolor[HTML]{F0E9FD} ImageNet \citep{deng2009imagenet} & 1,000 & 50,000 \\
\rowcolor[HTML]{F0E9FD} Caltech101 \citep{Caltech101} & 100 & 2,465 \\
\rowcolor[HTML]{F0E9FD} OxfordPets \citep{Pets} & 37 & 3,669 \\
\rowcolor[HTML]{F0E9FD} StanfordCars \citep{Cars} & 196 & 8,041 \\
\rowcolor[HTML]{F0E9FD} Flowers102 \citep{nilsback2008automated} & 102 & 2,463 \\
\rowcolor[HTML]{F0E9FD} Food101 \citep{food101} & 101 & 30,300 \\
\rowcolor[HTML]{F0E9FD} FGVCAircraft \citep{Aircraft} & 100 & 3,333 \\
\rowcolor[HTML]{F0E9FD} SUN397 \citep{SUN397} & 397 & 19,850 \\
\rowcolor[HTML]{F0E9FD} DTD \citep{DTD} & 47 & 1,692 \\
\rowcolor[HTML]{F0E9FD} EuroSAT \citep{helber2019eurosat} & 10 & 8,100 \\
\rowcolor[HTML]{F0E9FD}  UCF101 \citep{UCF101} & 101 & 3,783 \\
\hline 
ImageNet-A \citep{imagenetA} & 200 & 7,500 \\
ImageNetV2 \citep{imagenetV2} & 1,000 & 10,000 \\
ImageNet-R \citep{imagenetR} & 200 & 30,000 \\
ImageNet-Sketch \citep{wang2019learning} & 1000 & 50,889 \\
\hline
\end{tabular}
\label{appendix:table_dataset}
\end{table}

\begin{table*}[t]
\centering
\setlength{\abovecaptionskip}{0.1cm}
\setlength{\belowcaptionskip}{0.1cm}
\renewcommand\arraystretch{1.2}
\setlength\tabcolsep{8pt}
\caption{\textbf{Fine-Grained Classification}. Results for CLIP-ViT-B/16 are reported, providing the \textbf{Accuracy represented as Acc.($\uparrow$)} and \textbf{\ece ($\downarrow$)} metrics of the initialization, after applying \texttt{PromptAlign}, and after jointly employing
\texttt{PromptAlign} and our proposed \tca loss (please see main text for configuration details). Note that the baseline method  \texttt{PromptAlign} \cite{abdul2024align} is initialized with ‘a photo of a’ manual prompt. The values highlighted in \textbf{bold} indicate the lowest \ece achieved following test-time prompt tuning and \textbf{underline} is the second best. \textbf{Note:} The first 3 rows-pairs (\acc,\ece) are borrowed from \ctpt paper\cite{yoon2024ctpt}. We outperform \texttt{promptAlign}\cite{abdul2024align} and \ctpt\cite{yoon2024ctpt} both in terms of achieving lowest \ece and accuracy.}
\label{tab:suppl_tab_1}
\resizebox{\textwidth}{!}{%
\begin{tabular}{l|l|cccccccccc|c}
\hline
\textbf{Method} & \textbf{Metric} & \textbf{Caltech} & \textbf{Pets} & \textbf{Cars} & \textbf{Flower} & \textbf{Food101} & \textbf{Aircraft} & \textbf{SUN397} & \textbf{DTD} & \textbf{EuroSAT} & \textbf{UCF101} & \textbf{Average} \\ \hline\hline
\rowcolor[HTML]{FDFCEB} 
\cellcolor[HTML]{FDFCEB} & Acc. & 90.9& 82.5& 64.6& 64.7& 83.9& 22.3& 61.4& 42.4& 38.8& 64.8 & 61.63\\
\rowcolor[HTML]{FDFCEB} 
\multirow{-2}{*}{\cellcolor[HTML]{FDFCEB}CLIP-ViT-B/16\textsubscript{HardPrompt}} & ECE   &7.51& 2.91& \underline{2.49} & \underline{4.70} & 2.78& 7.09& 3.33& \underline{9.5}& 13.4& \textbf{2.79} & \underline{5.194}\\
\rowcolor[HTML]{FFF8F2} 
\cellcolor[HTML]{FFF8F2} & Acc. & 94.1& 90.5 &68.0& 72.1& 87.6 &25.5 &68.1 &47.9 &44.8& 69.8&66.84\\
\rowcolor[HTML]{FFF8F2} 
\multirow{-2}{*}{\cellcolor[HTML]{FFF8F2}+PromptAlign\textsubscript{HardPrompt}} & ECE  &2.30& 2.86& 1.98& 11.2& 3.04& 8.30 &8.39 &25.6 &24.7 &12.1 &10.04\\
\rowcolor[HTML]{FBEEEF} 
\cellcolor[HTML]{FBEEEF} & Acc. & 94.0& 90.6& 67.8 &72.1 &87.5& 25.3& 67.8& 47.7& 45.9& 69.8& 66.85  \\
\rowcolor[HTML]{FBEEEF} 
\multirow{-2}{*}{\cellcolor[HTML]{FBEEEF}+PromptAlign\textsubscript{HardPrompt}+C-TPT} & ECE&   \underline{2.20}& \textbf{2.09}& \textbf{1.79} &9.26 &\textbf{2.25} &6.57& 6.29 &22.1& 21.8&9.95&8.43\\
\rowcolor[HTML]{FFEFFC} 
\cellcolor[HTML]{FFEFFC} & Acc. & 93.31&	90.9&	65.84&	67.68&	86.28&	26.79&	66.78&	46.63&	44.06&	69.2& 65.75 \\
\rowcolor[HTML]{FFEFFC} 
\multirow{-2}{*}{\cellcolor[HTML]{FFEFFC}\textbf{+PromptAlign\textsubscript{HardPrompt}+TCA+2 Attributes}} & ECE & \textbf{2.17}&	\underline{4.21}&	6.85&	\textbf{4.41}&	\underline{2.86}&	\underline{2.5}&	\textbf{2.08}&	\textbf{9.45}&	\textbf{7.95}&	\underline{3.3}& \textbf{4.58} \\

\rowcolor[HTML]{F0E9FD} 
\cellcolor[HTML]{F0E9FD} & Acc.  & 93.06&	90.81&	66.01&	68.41&	86.61&	26.88&	67.48&	48.05&	45.93&	69.71& 66.30 \\
\hline

\end{tabular}%
}
\end{table*}

\section{Additional Experiments}
\label{sec:addlExp}

Tab. \ref{tab:suppl_tab_1} shows the results of the proposed method \tca in comparison with the contemporary methods. We add \texttt{PromptAlign} \cite{abdul2024align} individually and also combine \tca with \texttt{PromptAlign} as ablation. We outperform \texttt{PromptAlign}\cite{abdul2024align} both in terms of achieving the lowest \ece and accuracy. 




\begin{figure*}[t]
    \centering
    \vspace{3em}
    \includegraphics[width=1.0\linewidth]{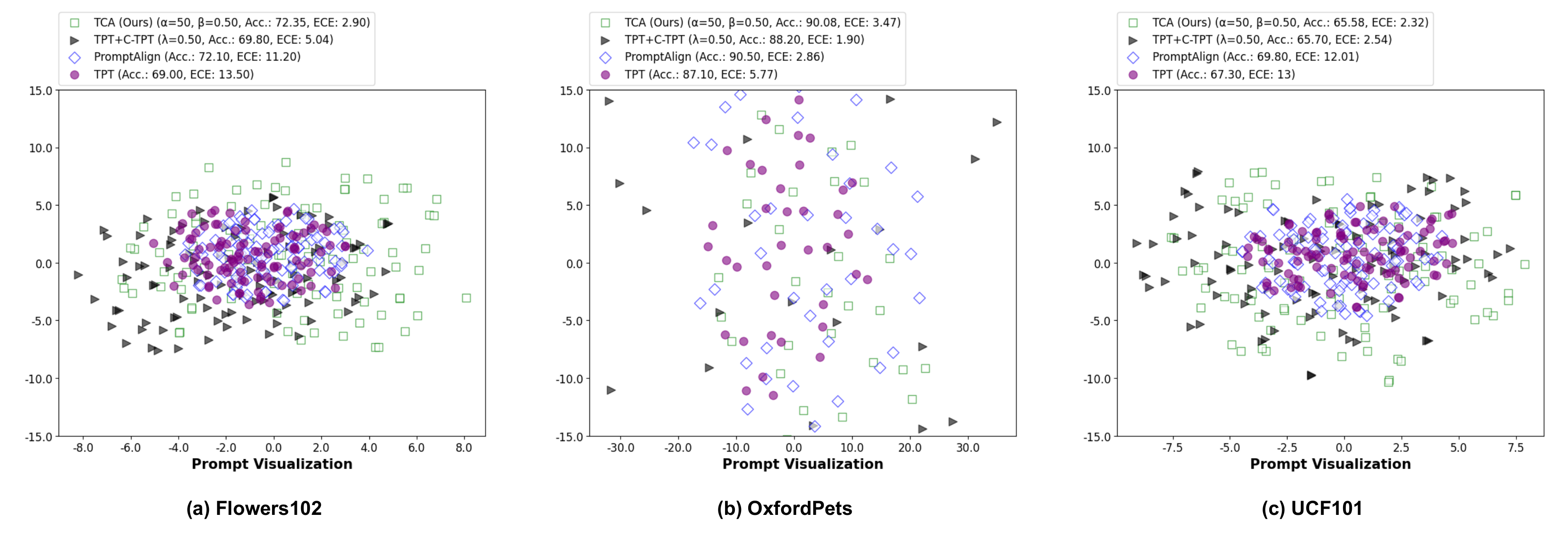} 
    
    \caption{\textbf{t-SNE visualization}: Class-specific Text Embeddings are shown via t-SNE for the tuned prompts on (a) \texttt{Flowers102} \citep{nilsback2008automated}, (b) \texttt{OxfordPets} \citep{Pets} and (c) \texttt{UCF101} \citep{UCF101} datasets. Each color in the figure denotes a unique prompt. We can see \tca exhibits the lowest \ece on \texttt{Flower102} \citep{nilsback2008automated} and \texttt{UCF101} \citep{UCF101} datasets, showing the maximum dispersion and hence better calibration. Experiments are with the VIT-B/16 model \cite{dosovitskiy2020image}.}
\label{fig:tsne_datasets}
\end{figure*}

\subsection{t-SNE Vizualization on additional datasets}

\begin{figure}[t]
	\begin{minipage}{\linewidth}
		\centering
    \includegraphics[width=\linewidth]{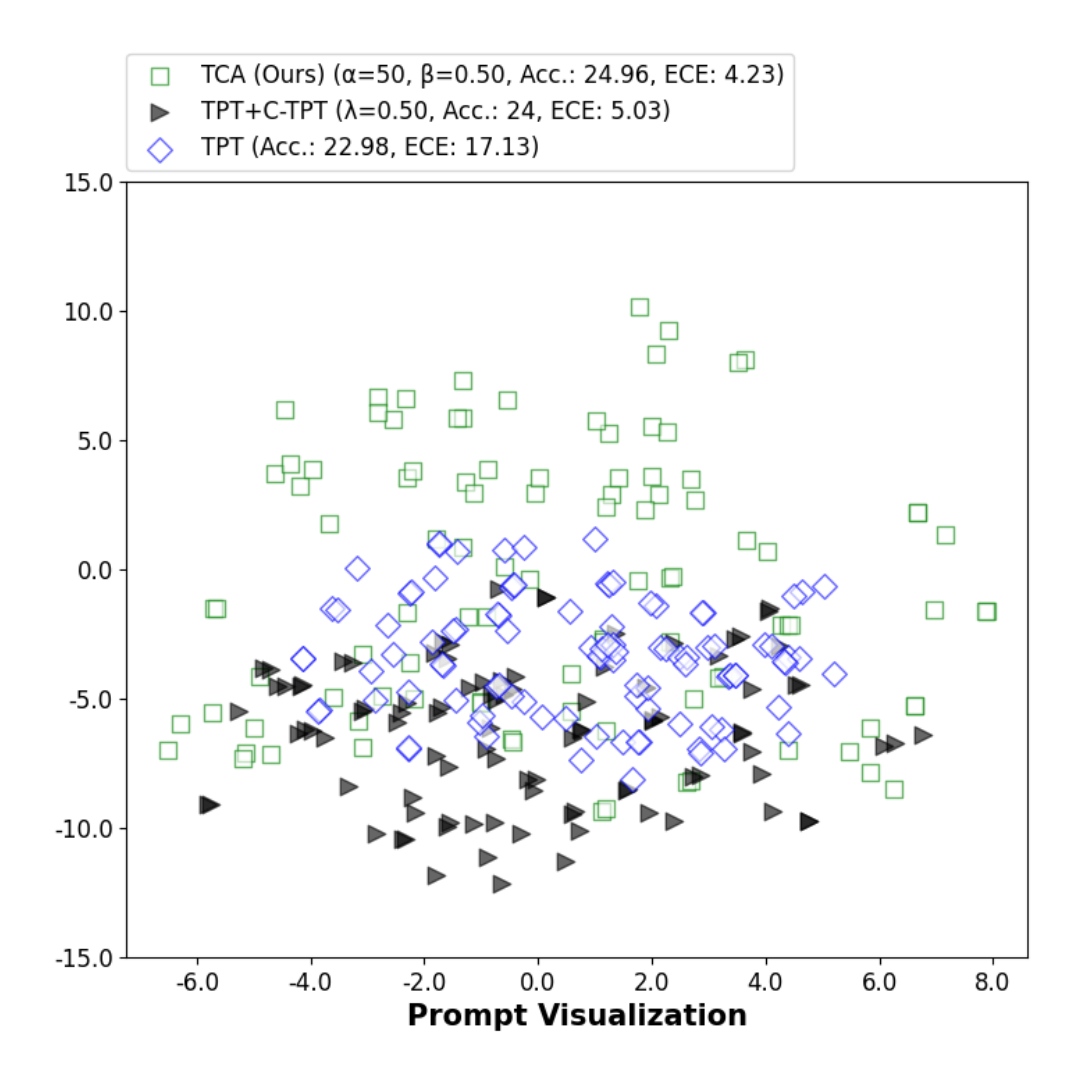}
    \end{minipage}
	\quad
	\begin{minipage}{\linewidth}
    \caption{The t-SNE plot of prompt visualizations for the proposed 
\tca is compared with the recent state-of-the-art method, \ctpt. It is observed that \tca demonstrates the highest class dispersion, indicating superior class separability.}
    \label{fig:tsneWithCTPT}
	\end{minipage}
\end{figure}

Figs. \ref{fig:tsneWithCTPT} and \ref{fig:dispersion} shows t-SNE \cite{tsne} plot to visualize the class-specific text embeddings of the tuned prompts, demonstrating varying levels of calibration. The result indicates that the prompts generated by methods like \tpt and \tpt+ individual terms of our loss (either intra or interclass losses) exhibit less dispersion and demonstrate lower calibration unlike our technique. 
\tca surpasses in improving calibration by strategically enhancing the diversity of text features through targeted attribute application. For additional t-SNE plots on different datasets and contemporary method, please see the supplementary material.

We present additional t-SNE plots illustrating the visualization of class-specific text embeddings generated by the tuned prompts across three datasets: (a) Flowers102 \citep{nilsback2008automated}, (b) OxfordPets \citep{Pets}, and (c) UCF101 \citep{UCF101}, as shown in \cref{fig:tsne_datasets}. For both the Flowers102 \citep{nilsback2008automated} and UCF101 \citep{UCF101} datasets, the \tca-tuned prompts demonstrate better calibration, characterized by a more dispersed cluster. In contrast, on the OxfordPets \citep{Pets} dataset, \tpt+\ctpt performs better, resulting in more dispersed tuned prompts, indicative of an improved embedding separation.

\begin{table*}[t]
\centering
\setlength{\abovecaptionskip}{0.1cm}
\setlength{\belowcaptionskip}{0.1cm}
\renewcommand\arraystretch{1.2}
\setlength\tabcolsep{8pt}
\caption{\textbf{Fine-Grained Classification using DiffTPT}: Results for CLIP-RN50 and CLIP-ViT-B/16 are reported, providing the Accuracy represented as Acc.(↑) and ECE (↓) metrics of the initialization, after applying \texttt{DiffTPT} \cite{feng2023diverse}, and after jointly employing \texttt{DiffTPT} and our proposed \texttt{TCA} loss (please see main text for configuration details). Note that the baseline method \texttt{DiffTPT} is initialized with ‘a photo of a’ manual prompt. The values highlighted in bold indicate the lowest ECE achieved following test-time prompt tuning and underline is the second best. We outperform \texttt{DiffTPT} both in terms of achieving lowest ECE on an average.}
\label{tab:suppl_tab_2}
\resizebox{\textwidth}{!}{%
\begin{tabular}{l|l|llllllllll|l}
\hline
\textbf{Method}                                                                & \textbf{Metric} & \textbf{Caltech} & \textbf{Pets} & \textbf{Cars} & \textbf{Flower} & \textbf{Food101} & \textbf{Aircraft} & \textbf{SUN397} & \textbf{DTD} & \textbf{EuroSAT} & \textbf{UCF101} & \textbf{Average} \\ \hline \hline
\rowcolor[HTML]{FBEEEF} 
\cellcolor[HTML]{FBEEEF}                                                       & Acc.            & 86.21            & 83.64         & 60.2          & 63.78           & 79.23            & 17.84             & 62.11           & 40.88        & 41.36            & 62.41           & 59.76           \\
\rowcolor[HTML]{FBEEEF} 
\multirow{-2}{*}{\cellcolor[HTML]{FBEEEF}DiffTPT: CLIPRN50}                    & ECE             &  4.83                & 	6.37              &  	\textbf{4.11}             & 	7.71                &  	4.15                & 6.89                  & 	\textbf{3.73}                & 	10.12             & 	16.37                 & 	\textbf{3.54}                &  	6.78                \\
\rowcolor[HTML]{FFEFFC} 
\cellcolor[HTML]{FFEFFC}                                                       & Acc.            &  87.44                &  	84.21             &  	60.95	             & 64.82	                & 80.11                  &  	17.91                 & 	62.64                & 	41.83             & 	41.11                 & 	62.81                &   	60.38               \\
\rowcolor[HTML]{FFEFFC} 
\multirow{-2}{*}{\cellcolor[HTML]{FFEFFC}\textbf{DiffTPT + TCA (2-attribute)}} & ECE             &  \textbf{4.1}                &  	\textbf{5.12}             &  	4.63             & 	\textbf{5.21}                & 	\textbf{3.86}                 &   	\textbf{4.62}                & 	\underline{4.56}                & 	\textbf{9.13}             & 	\textbf{12.28}                 & 	\underline{3.7}                &   	\textbf{5.72}               \\
\rowcolor[HTML]{F0E9FD} 
\cellcolor[HTML]{F0E9FD}                                                       & Acc.            &  87.63	                & 84.63	              &  59.96             &  	65.62	               & 80.27                 &  	17.81                 &  	62.83               &  	42.07            & 	41.41	                 &  63.02	               &  60.53                \\
\rowcolor[HTML]{FBEEEF} 
\cellcolor[HTML]{FBEEEF}                                                       & Acc.            & 92.32            & 88.39         & 67.33         & 70.01           & 87               & 25.02             & 65.89           & 47.12        & 43.83            & 68.43           & 65.53          \\
\rowcolor[HTML]{FBEEEF} 
\multirow{-2}{*}{\cellcolor[HTML]{FBEEEF}DiffTPT: ViT-B/16}                    & ECE             &                  2.73&               \textbf{2.75}&               \textbf{1.78}&                 9.68&                  3.41&                   9.23&                 7.73&              24.59&                  23.14&                 11.74&                  9.68\\
\rowcolor[HTML]{FFEFFC} 
\cellcolor[HTML]{FFEFFC}                                                       & Acc.            & 92.36            & 88.43         & 67.1          & 68.41           & 86.94            & 25.65             & 65.64           & 47.42        & 42.1             & 68.34           & 65.24           \\
\rowcolor[HTML]{FFEFFC} 
\multirow{-2}{*}{\cellcolor[HTML]{FFEFFC}\textbf{DiffTPT + TCA (2-attribute)}} & ECE             &                  \textbf{2.65} &               4.78&               \underline{6.22}&                 \textbf{4.64}&                  \textbf{3.07}&                   \textbf{3.57}&                 \textbf{3.01}&              \textbf{10.02}&                  \textbf{8.03}&                 \textbf{3.88}&                  \textbf{4.99}\\
\hline
\end{tabular}}
\end{table*}

\begin{table*}[t]
\centering
\setlength{\abovecaptionskip}{0.1cm}
\setlength{\belowcaptionskip}{0.1cm}
\renewcommand\arraystretch{1.2}
\setlength\tabcolsep{8pt}

\caption{\textbf{Natural Distribution Shifts for DiffTPT}. Results for CLIP-RN50 and CLIP-ViT-B/16 are reported for DiffTPT \cite{feng2023diverse}, providing the \textbf{Acc. ($\uparrow$)} and \textbf{ECE ($\downarrow$)} metrics for different experimental configurations (please refer to the main text for details of configurations). Dataset abbreviations: ImageNet-V2 (IN-V2), ImageNet-A (IN-A), ImageNet-R (IN-R), and ImageNet-Sketch (IN-S). Values highlighted in \textbf{bold} indicate the lowest \ece achieved after test-time prompt tuning.}
\label{tab:suppl_tab_2}

\resizebox{1.0\textwidth}{!}{%
\begin{tabular}{l|l|llll|l}
\hline
\textbf{Method}                                                                & \textbf{Metric} & \textbf{IN-A} & \textbf{IN-V2} & \textbf{IN-R} & \textbf{IN-S} & \textbf{Avg.}                 \\ \hline \hline
\rowcolor[HTML]{FBEEEF} 
\cellcolor[HTML]{FBEEEF}                                                       & Acc.            & 31.51         & 55.56          & 58.8          & 37.1          & 46                            \\
\rowcolor[HTML]{FBEEEF} 
\multirow{-2}{*}{\cellcolor[HTML]{FBEEEF}DiffTPT: CLIP-RN50}                   & ECE             & 19.76              & 14.43               & 	8.21              & 	17.89              & 	15.07                              \\
\rowcolor[HTML]{FFEFFC} 
\cellcolor[HTML]{FFEFFC}                                                       & Acc.            &  31.07             & 	55.79	               & 57.1              & 	37.03               & 	45.25                              \\
\rowcolor[HTML]{FFEFFC} 
\multirow{-2}{*}{\cellcolor[HTML]{FFEFFC}\textbf{DiffTPT + TCA (2-attribute)}} & ECE             & \textbf{18.47}              & 	\textbf{7.87}               & 	\textbf{7.67}              & 	\textbf{9.29}              & 	\textbf{10.83}                              \\
\rowcolor[HTML]{FBEEEF} 
\cellcolor[HTML]{FBEEEF}                                                       & Acc.            & 55.81         & 65.34          & 75            & 46.8          & 60.74                         \\
\rowcolor[HTML]{FBEEEF} 
\multirow{-2}{*}{\cellcolor[HTML]{FBEEEF}DiffTPT: CLIP-ViT-B/16}               & ECE             &               13.56&                12.14&               \textbf{5.23}&               14.67&                               11.4\\
\rowcolor[HTML]{FFEFFC} 
\cellcolor[HTML]{FFEFFC}                                                       & Acc.            & 52.37         & 62.76          & 73.56         & 45.3          & \cellcolor[HTML]{FBEEEF}58.5  \\
\rowcolor[HTML]{FFEFFC} 
\multirow{-2}{*}{\cellcolor[HTML]{FFEFFC}\textbf{DiffTPT + TCA (2-attribute)}} & ECE             &               \textbf{4.67}&                \textbf{2.89}&               \underline{6.11}&               \textbf{3.47}& \cellcolor[HTML]{FBEEEF}      \textbf{4.28}\\
\hline
\end{tabular}}
\end{table*}

\end{document}